%% file: main.tex
\documentclass{article} 
\usepackage[final]{colm2025_conference}

\usepackage{microtype}
\usepackage{hyperref}
\usepackage{url}
\usepackage{booktabs}
\usepackage{graphicx}
\usepackage{adjustbox}

\usepackage{fontawesome5}

\usepackage{lineno}

\usepackage{wrapfig}

\usepackage{graphicx}
\usepackage{changepage}
\usepackage{longtable}
\usepackage{svg}

\usepackage{titlesec} 
\usepackage{hyperref} 
\usepackage{tabularx}
\usepackage{enumitem}
\usepackage{xcolor}
\usepackage[most]{tcolorbox} 
\usepackage{longtable}
\lstdefinelanguage{Markdown}{
  basicstyle=\ttfamily\footnotesize,
  sensitive=false,
  morecomment=[l]{\#},   
  morecomment=[s]{```}{```},
  morestring=[b]",        
  morestring=[b]', 
}
\usepackage{inconsolata}

\PassOptionsToPackage{table}{xcolor}
\usepackage{colortbl}
\usepackage[table]{xcolor}

\definecolor{green}{RGB}{152, 203, 177}

\newcommand{\levelcolor}[1]{%
    \pgfmathsetmacro{\opacity}{(#1-10)/(77-10)}%
    \cellcolor{green!\number\numexpr\opacity*100\relax!white}%
    #1\%%
}

\newcolumntype{R}{>{\collectcell\levelcolor}c<{\endcollectcell}}

\definecolor{lightcayenne}{HTML}{b6122c}
\setlist{nolistsep}
\definecolor{orange}{rgb}{0.85, 0.33, 0.2}

\newcommand{\naive}{\texttt{NA\"IVE}}
\newcommand{\pipeline}{\texttt{CLIPPER}}

\newcommand{\llamainst}{Llama-Instruct}
\newcommand{\llamaft}{Llama-\pipeline}
\newcommand{\llamaftbalanced}{Llama-\pipeline}

\newcommand{\qweninst}{Qwen-Instruct}
\newcommand{\qwenft}{Qwen-\pipeline}
\newcommand{\qwenftbalanced}{Qwen-\pipeline}

\newcommand{\prolongbase}{ProLong-Base}
\newcommand{\prolonginst}{ProLong-Instruct}
\newcommand{\prolongwp}{ProLong-WritingPrompts}

\newcommand{\prolongftbalanced}{ProLong-\pipeline}
\newcommand{\prolongftbook}{ProLong-\pipeline-book}
\newcommand{\prolongftchapter}{ProLong-\pipeline-chapter}

\definecolor{darkblue}{rgb}{0, 0, 0.5}
\hypersetup{colorlinks=true, citecolor=darkblue, linkcolor=darkblue, urlcolor=darkblue}
\definecolor{lowgreen}{RGB}{200, 255, 200}   
\definecolor{highgreen}{RGB}{0, 150, 0}      

\title{
\includegraphics[width=0.55cm]{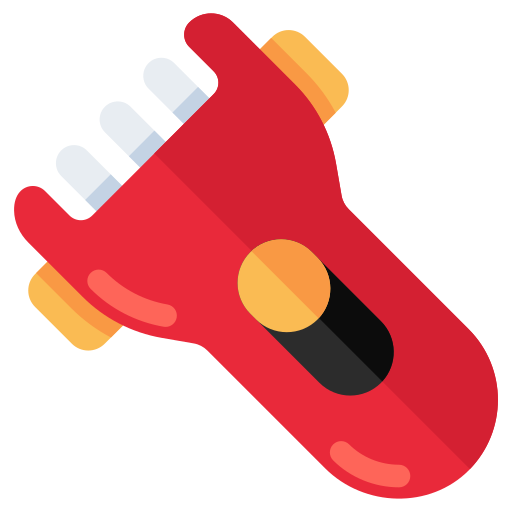}\textsc{Clipper}: Compression enables long-context synthetic data generation}

\author{
  Chau Minh Pham\textsuperscript{1} \quad
  Yapei Chang\textsuperscript{1} \quad
  Mohit Iyyer\textsuperscript{1,2}\\[0.5em]
  \textsuperscript{1}University of Maryland, College Park \quad
  \textsuperscript{2}University of Massachusetts Amherst\\[0.5em]
  \texttt{\{chau,yapeic,miyyer\}@umd.edu}
}

\begin{document}

\ifcolmsubmission
\linenumbers
\fi

\maketitle

\input{sections/0-abstract}
\input{sections/1-intro}
\input{sections/2-data}
\input{sections/3-experiments}
\input{sections/4-results}
\input{sections/5-related}
\input{sections/6-conclusion}
\input{sections/7-limitations-ethics}
\section*{Acknowledgment}

We extend special gratitude to Marzena Karpinska for benchmarking our models on NoCha and giving us helpful insights into the models' behaviors. This project was partially supported by awards IIS-2046248, IIS-2312949, and IIS-2202506 from the National Science Foundation (NSF). 

\bibliography{references,custom}
\bibliographystyle{colm2025_conference}
\input{sections/A-data}
\input{sections/A-training}

\input{sections/A-evaluation}
\input{sections/A-results}

\end{document}

%% file: sections/0-abstract.tex
\begin{abstract}
    LLM developers are increasingly reliant on synthetic data, but generating high-quality data for complex long-context reasoning tasks remains challenging. We introduce \pipeline,\footnote{\pipeline\ stands for \textbf{C}ompressing \textbf{L}ong \textbf{I}n\textbf{P}uts.} a compression-based approach for generating synthetic data tailored to \textit{narrative claim verification} -- a task that requires reasoning over a book to verify a given claim. Instead of generating claims directly from the raw text of the book, which results in artifact-riddled claims, \pipeline\ first compresses the book into chapter outlines and book summaries and then uses these intermediate representations to generate complex claims and corresponding chain-of-thoughts. Compared to na\"ive approaches, \pipeline\ produces claims that are more valid, grounded, and complex. Using \pipeline, we synthesize a dataset of 19K claims paired with source books and chain-of-thought reasoning, and use it to fine-tune three open-weight models. Our best model achieves breakthrough results on narrative claim verification (from 28\% to 76\% accuracy on our test set) and sets a new state-of-the-art for sub-10B models on the NoCha leaderboard. Further analysis shows that our models generate more detailed and grounded chain-of-thought reasoning while also improving performance on other narrative understanding tasks (e.g., NarrativeQA).
    \begin{center}
        \faGithub[regular]  {\small \url{https://github.com/chtmp223/CLIPPER}}
    \end{center}
\end{abstract}

%% file: sections/1-intro.tex
\section{Introduction}

Due to the high cost of human-annotated data, LLM developers increasingly rely on \emph{synthetic} data (generated by LLMs) to boost instruction following and reasoning capabilities~(\citealt{ding_enhancing_2023, lambert2024tulu3, yang_qwen25-1m_2025}, \emph{inter alia}). 
As the context size of LLMs extends to millions of tokens, it is important to ensure that we have scalable and performant strategies to create synthetic data for \emph{long-context} tasks.
Prior work creates such data by (1) selecting a long document or a smaller chunk within; and (2) prompting an LLM to generate input/output pairs using the selected text \citep{bai_longalign_2024, dubey2024llama}. 

While this strategy is effective for tasks like summarization and QA, we show that it breaks down for more complex reasoning-oriented tasks like \emph{narrative claim verification}, in which a model must judge whether a statement about a long input text is true or false.
The majority of narrative claims in the NoCha benchmark~\citep{karpinska_one_2024}, which was created by human readers of fictional books, can only be verified by \emph{global} reasoning over events, characters, and relationships.\footnote{Creating NoCha is costly and challenging: annotators read full books and earn \$1.70 per claim.} This poses a challenge to even the best LLMs: OpenAI's o1-preview currently leads with an accuracy of 67.4\% (far below human performance). If no LLM can reliably solve the task, how can we produce and validate synthetic data for it? 

\begin{figure*}[!htbp]
    \centering
    \includegraphics[width=\linewidth]{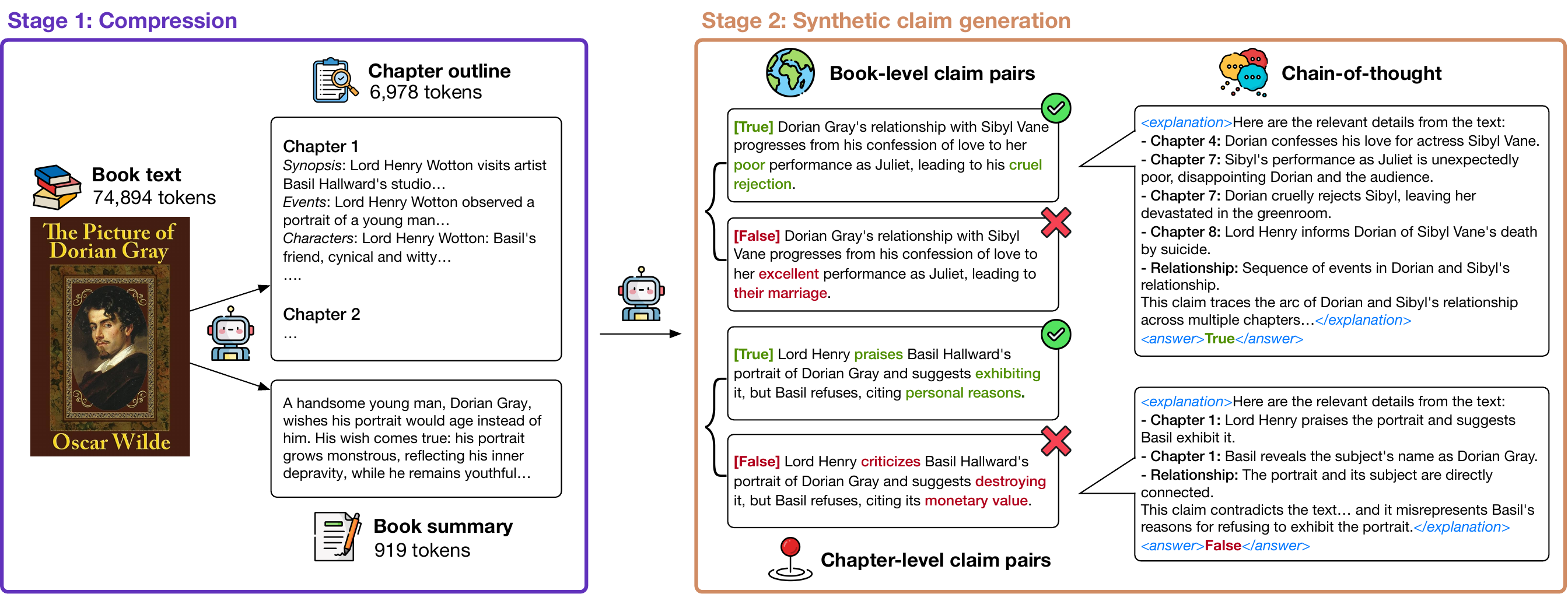}
    \caption{\pipeline\ overview. (1) \textit{Compression}: An LLM generates chapter outlines (8.7K tokens) and summaries (618 tokens) from books (90K tokens). (2) \textit{Claim generation}: The LLM produces true/false claims with chains-of-thought with the outlines and summaries.
    }
    \label{fig:overview}
    \vspace{-2em}
\end{figure*}

We tackle this challenge by introducing \pipeline, a synthetic data generation pipeline that operates in two stages (Figure \ref{fig:overview}). First, a long document is \emph{compressed} by an LLM into summaries and/or outlines that contain salient events. Then, the LLM generates claims based on the compressed narrative, with optional instructions to write claims that require reasoning over multiple chapters to verify. Each claim is accompanied by a chain-of-thought reasoning trace that grounds it within specific chapters where relevant events occur.

Compared to the na\"ive strategy of prompting an LLM with the entire (uncompressed) book, \pipeline\ significantly reduces the error rate in the claims from 73.1\% to 16.7\%, while also producing more claims at a lower cost. Why does this work? Prior work has shown that LLMs are high-quality summarizers of long documents~\citep{chang_booookscore_2024, kim_fables_2024}. By operating on compressed representations, we also address the known degradation of instruction-following in long-context settings~\citep{wu_lifbench_2024, levy_same_2024} and thus reduce the complexity of the claim generation process.

We use \pipeline\ to generate a dataset containing 19K claims about public-domain fictional books. Fine-tuning open-weight models like Llama-3.1-8B-Instruct \citep{dubey2024llama}, ProLong-512K-8B-Base \citep{gao_how_2024} and Qwen2.5-7B-Instruct~\citep{qwen_qwen25_2024} on this dataset yields large improvements on narrative claim verification and positive transfer to other narrative-related tasks (e.g., NarrativeQA, MuSR). For instance, fine-tuning Llama-3.1-8B-Instruct on our data doubles its NoCha performance (from 16.5\% to 32.2\%) and almost triples its test set performance (from 27.9\% to 76.0\%). Our fine-tuned Qwen model sets a new state of the art on NoCha for $<$10B models, outperforming closed models like Gemini 1.5 Flash 8B~\citep{geminiteam2024gemini15unlockingmultimodal} and OpenAI o1-mini~\citep{openai_o1_2024}.


While our approach is promising for improving long-context reasoning in open-weight models, our best models fall well short of the performance reached by closed LLMs such as o1 on NoCha. The performance gap between NoCha and \pipeline-test\ likely stems from the nature of the claims, as \pipeline-test\ features synthetic claims based on model-generated outlines, while NoCha’s human-written claims require reasoning about details often absent from such outlines. We analyze where our fine-tuned models can still improve, discovering that they benefit more from training on claims whose evidence is localized to a single chapter (and not more complex multi-chapter claims). We hope future work will use our data generation pipeline for fine-tuning larger models (e.g., $>$70B) on other long-context tasks to further improve global reasoning abilities. In summary, our contributions are:
\vspace{-0.5em}
\begin{enumerate}
    \item \pipeline: A compression-based pipeline for synthesizing grounded long-context data at low cost, producing 19K claim-book pairs for narrative claim verification.
    \item Fine-tuned open-weight models that advance claim verification and narrative understanding, showing the benefits of training on data generated with \pipeline.
\end{enumerate}

%% file: sections/2-data.tex
\section{\pipeline: generating high-quality synthetic data via compression}
In long-context settings, synthetic datasets have typically been created by selecting lengthy documents from an existing corpus and using an LLM to generate input-output pairs given either the entire document~\citep{bai_longalign_2024} or random excerpts~\citep{dubey2024llama, yang_qwen25-1m_2025}.
For our task, however, we show that these methods are insufficient:
\begin{itemize}
\item \textit{Providing the LLM with the entire document}  results in much noisier data, as we show that producing high-quality, complex claims about long narratives is a fundamentally difficult task even for the best models (\S\ref{data:alt}).
\item \textit{Providing only an excerpt from the long document}, on the other hand, precludes the model from generating claims that require global reasoning across the entire book.
\end{itemize}
We thus develop a two-stage strategy, \pipeline, which first compresses the narrative into chapter outlines and summaries, then prompts an LLM to produce claims and chain-of-thoughts grounded in the compressed narrative (\S\ref{data:claims_generation}).
We use the dataset generated by \pipeline\ to fine-tune open-weight models in \S\ref{sec:training}. \textbf{All prompts can be found in \S\ref{appendix:data-construction}.}


\subsection{Task setup}

Before describing how \pipeline\ works, we first establish the definition for \textit{narrative claim verification}, then explain how we collect the books that serve as the foundation for this task.

\paragraph{Task definition:}
In \textit{narrative claim verification}, an LLM is given a book and a claim about the book. The task is to determine whether the claim is true or false while providing a clear explanation for its decision. A key aspect of the task is the inclusion of \textbf{true/false} \textit{narrative minimal pairs}~\citep{karpinska_one_2024}, where each false claim closely resembles its true claim counterpart but contains subtle inaccuracies (illustrated in Figure \ref{fig:overview}; see Table \ref{tab:error-analysis-dist} for more examples). The model is considered accurate only if it correctly verifies both claims in a pair, which reduces the chances of the model being correct for the wrong reason.

\paragraph{Gathering public domain books:} We collect \textbf{479} fictional books from Project Gutenberg, with an average length of \textbf{90K} tokens\footnote{All token counts are computed using o200k\_base from \url{https://github.com/openai/tiktoken}.} and \textbf{23} chapters.\footnote{We do not include books longer than 128K tokens as many open-weight models cannot process anything beyond that number. We clean the manually downloaded books by removing supplementary content to prevent models from using these metadata as shortcuts for event retrieval.} While these texts might raise memorization concerns, \S\ref{appendix:memorization} shows they do not affect baseline model performance.\footnote{\href{https://chatgptiseatingtheworld.com/wp-content/uploads/2025/01/Unredacted-Reply-of-Plaintiffs-1.pdf}{https://chatgptiseatingtheworld.com/wp-content/uploads/2025/01/Unredacted-Reply-of-Plaintiffs-1.pdf} shows that Llama models might have been trained on LibGen book data.}



\subsection{Na\"ive claim generation using book texts}
\label{data:alt}
One simple data synthesis approach is prompting an LLM to generate claims directly from the book text. However, this \naive\ method falls short in a long-context setting, with \textbf{73.1\%} of the generated claims containing serious errors. In addition, \naive\ produces fewer claims at a higher price. These limitations motivate us to develop \pipeline.

\paragraph{The \naive\ method:} We provide Claude-3.5-Sonnet-v1 with the entire book text and prompt it to generate pairs of true/false claims along with corresponding chain-of-thought reasoning in a zero-shot manner. Note that we cannot use few-shot prompting due to the books' length. We finally prompt Claude to remove any duplicated claims among the generated claims. Note that we do not ask Claude to validate the generated claims, as this is a much more challenging task (as seen by Claude's 40.3\% accuracy on NoCha)—the very problem we aim to address in this paper. In contrast, claim validation is very feasible with \pipeline, as Claude's claim verification accuracy given outlines instead of text is 98.6\%. 

\paragraph{Human validation of \naive\ claims:} We manually annotate 52 claims generated by \naive\ based on six books (Table~\ref{tab:six-books}).\footnote{We annotate each claim with its most major error type.}
Across these claims, we identify four types of errors. Table~\ref{tab:naive} shows that 73.1\% of \naive\ claims contain an error. Specifically, there are 11.5\% \textit{invalid} claims, often due to mislabeled false claims that are actually valid. There is also a high number of \textit{misattributed} (28.9\%) claims that cite the wrong chapters in the produced chain-of-thought. 17.3\% of the claims are \textit{duplicated} despite the deduplication step, because Claude frequently hallucinates source chapters. 15.4\% of the claims also include \textit{explicit references} to chapter numbers or direct quotes,\footnote{This happens despite explicit formatting instructions.} which compromises the task by revealing the evidence location. Additionally, we observe that the events referenced in the generated claims are often the book's most major events, making the claims much easier for LLMs to verify. Beyond these errors, the \naive\ pipeline is costly at $\approx$\$0.07 USD per claim ($\approx$ \$1,330 for 19K claims).

\begin{table*}[t]
\centering
\footnotesize
\resizebox{\textwidth}{!}{%
\begin{tabular}{p{0.09\textwidth}p{0.005\textwidth}p{0.005\textwidth}p{0.25\textwidth}p{0.66\textwidth}}
\toprule
\multicolumn{1}{c}{\textsc{Category}} & \multicolumn{1}{c}{\naive} & \multicolumn{1}{c}{\pipeline} &\multicolumn{1}{c}{\textsc{Error Definition}} & \multicolumn{1}{c}{\textsc{Example}} \\
\midrule
Invalid & 11.5\% &9.1\% & The claim is incorrect with respect to the book text, or the true/false claim pair is invalid. & Anne rejects three marriage proposals during her time at Redmond College: from Charlie Sloane, Gilbert Blythe, and Roy Gardner, {\color{purple}{all because she doesn't love them}}. \textit{(This false claim is not entirely false because Anne really didn’t love them or wasn’t initially aware of her romantic feelings.)}. \\
\midrule
Mis-attribution & 28.9\%& 4.6\%& The claim is valid, but the associated explanation does not cite the correct chapters. & Dr. Sheppard...was the last person known to have seen Roger Ackroyd alive at 8:50 PM on the night of the murder, and he later assisted Hercule Poirot in the investigation while simultaneously {\color{purple}{concealing Ralph Paton in a nursing home}}. \textit{(The explanation cites Chapter 1, 4, 16, and 20, but misses Chapter 24, which mentions that Ralph is in a nursing home)}\\
\midrule
Explicit references & 15.4\% & 0.0\%& The claim is easier to verify since it includes direct quotes and chapter references, eliminating the need for event retrieval. & Alice's pursuit of the White Rabbit, which begins with her following him down a rabbit hole in {\color{purple}{Chapter 1}}, continues throughout her adventure, including an encounter in the King and Queen of Hearts' court in {\color{purple}{Chapter 11}} where the Rabbit acts as a herald.\\
\midrule
Duplication & 17.3\%& 3.0\%& The claim describes the same events as another. Although their content is similar, differences in wording may allow both to pass our deduplication process.& "Dorian Gray's cruel rejection of Sibyl Vane after her poor performance as Juliet leads to her suicide, which Dorian callously dismisses by attending the opera the following night, resulting in the first noticeable change in his portrait [...]" versus "Dorian Gray's cruel rejection of Sibyl Vane after her poor performance as Juliet leads to her suicide, causing the first visible change in his portrait [...] culminate in his murder of Basil Hallward years later [...]." \\
\midrule
\textbf{Any error} & \textbf{73.1\%} & \textbf{16.7\%} & \multicolumn{2}{l}{The claim is invalid, misattributed, duplicated, or contains explicit references.} \\
\bottomrule
\end{tabular}
}
\caption{Error types among claims produced by \naive\ (52) and \pipeline\ (66) based on six books from Table \ref{tab:six-books}. Examples are selected from \naive\ claims.
}
\label{tab:naive}
\end{table*}

\subsection{Claim generation with \pipeline} 
\label{data:claims_generation}
To produce more valid and grounded claims, we use \textit{compressed} representations of the book content, namely chapter outlines and book summaries. These intermediate forms help anchor claims to specific events in the book, reducing the need to search through the entire text for relevant details. Additionally, this approach makes it easier to generate claims about lower-level events, addressing a major limitation of the \naive\ approach. We first (1) compress the books into a chapter outline and book summary and then (2) generate pairs of true/false claims at different scopes based on these compressed representations.

\paragraph{(1) Compressing books into summaries and chapter outlines:} Book summaries provide a global context for claim generation to ensure that each claim is consistent with the entire book. We prompt GPT-4o to summarize the entire book into a few paragraphs ($\approx$ 618 tokens on average).
Chapter outlines provide a list of fine-grained events that can be used to construct grounded claims. We prompt Claude\footnote{We set temperature=0.0, max\_tokens=4096. We use Claude instead of GPT-4o because Claude includes more concrete and objective events for the outline.} with each chapter text to generate an outline containing a synopsis, major events (5–7 per chapter), and a character list. Our compression rate is 10.0\%, calculated by averaging the ratio of outline length (8,745 tokens on average) to full book length (90,437 tokens) across all books.

\paragraph{(2a) Generating claims from compressed narratives:} We use chapter outlines and book summaries to generate true/false claims. We synthesize claims at two different scopes to enable reasoning across different token ranges: 

\noindent \textcolor{orange}{\textbf{$>$ Book-level claims:}} Claude is prompted to identify 2–3 key events from the outlines of at least 2 chapters, then use them to generate a claim. These claims require models to have a global understanding spanning multiple parts of the book.

\noindent \textcolor{orange}{\textbf{$>$ Chapter-level claims:}} Given the book summary and a single chapter outline, Claude is instructed to identify 2-3 key events of the chapter and use them to write a claim. While these claims do not necessitate global reasoning, they still require the model to search for correct chapter within a long text and perform intra-chapter reasoning (\S\ref{subsection:chap-book-ft}).



\paragraph{(2b) Deduplicating and validating generated claims:} Just as in \naive, we use Claude to remove duplicate claims. Additionally, we use GPT-4o\footnote{Chosen to mitigate potential self-biases \citep{xu2024prideprejudicellmamplifies, panickssery2024llmevaluatorsrecognizefavor, li2025preferenceleakagecontaminationproblem}.} to validate the claims against the source chapter outlines by prompting it to assess whether all parts of a claim are supported by the outline. This step is another advantage of \pipeline: unlike the \naive\ approach, our method allows for claim verification using the compressed chapter outline. To evaluate the reliability of LLM-based filtering, we manually review 72 claim pairs
and only disagree in one instance, where GPT-4o deems a claim valid that we find too subjective. Overall, 59.4\% of the original claims are removed as duplicates, while 2.4\% are filtered out as invalid.

\paragraph{Human validation of \pipeline\ claims:}

We use the same setup as described in \S\ref{data:alt} to manually evaluate 66 claims generated by the \pipeline\ pipeline. Table~\ref{tab:naive} provides a detailed breakdown of issues flagged in these claims, such as explicit references, invalidity, duplication, or misattribution. Notably, \textbf{83.3\% of the 66 claims are found to be completely free of errors—a significant improvement compared \naive's 26.9\%}. \pipeline\ also costs less at \$0.05 USD per claim compared to \naive\ (\$0.07 per claim) and human annotators (\$1.7 per claim based on \citealt{karpinska_one_2024}).\footnote{See detailed cost analysis in \S\ref{appendix:data-cost}.} 

\subsection{Automatic validation of \pipeline\ chain-of-thoughts}
\label{data:cot_validation}
We assess the groundedness of chain-of-thought (CoT) reasoning by prompting an LLM to verify whether each event in the CoT is supported by the chapter outline. Accuracy is measured as the percentage of events in true claim CoTs that are grounded in the book. To scale up evaluation, we use an LLM judge, DeepSeek-R1-Distill-Llama-70B~\citep{deepseekai2025deepseekr1incentivizingreasoningcapability}. 
We find that 98.5\% of CoTs are grounded. The remaining ungrounded CoTs typically involve events open to multiple interpretations. Compared to \naive, \pipeline's CoTs are significantly easier to verify due to their explicit chapter references.

%% file: sections/3-experiments.tex
\section{Supervised fine-tuning for LLMs on \pipeline\ data}
\label{sec:training}
Having shown that \pipeline\ produces synthetic data of high quality, we now investigate the effects of training on such data. We apply supervised fine-tuning (SFT) to three models on our dataset: ProLong-512K-8B-Base \citep{gao2024trainlongcontextlanguagemodels},\footnote{Despite the name, this model has undergone instruction tuning before. The ProLong team ran continual pre-training on Llama-3-8B-Instruct to get this model.} Llama-3.1-8B-Instruct \citep{dubey2024llama}, and Qwen2.5-7B-Instruct \citep{qwen2.5}.\footnote{We will now refer to these models as \prolongbase, \llamainst, and \qweninst.} Our top model, \llamaft, achieves nearly three times the test set performance of \llamainst—boosting accuracy from 27.9\% to 76\%—while showing substantial gains in long-context reasoning and narrative understanding on tasks like NoCha, NarrativeQA, and MuSR. Moreover, all of our models outperform all existing $<$10B models on the NoCha benchmark.

\subsection{Training setup}

\paragraph{Data splits and hyperparameters:} We divide our dataset into three parts: 16K claims (8K true/false pairs) for training, 2K for validation, and 1K for testing. Notably, the books in the test set do not overlap with those in the training or validation sets. For each entry, we combine the book text and claim to form the user prompt, and include the chain of thought reasoning along with the final answer as the assistant’s message (see Figure \ref{fig:sft_prompt}). A learning rate of 1e-6 and a batch size of 16 yield the best performance on our dev set.\footnote{We perform hyperparameter tuning on learning rates of 1e-5, 1e-6, and 1e-7, along with batch sizes of 16 and 32. Tuning is done for one epoch on a subset of 2K training samples. Due to high GPU costs (each epoch takes $~$5 hours), we only conduct hyperparameter tuning on \prolongbase\ only.} We fine-tune \qweninst, \llamainst, and \prolongbase\ using this configuration for one epoch.


\paragraph{Ablation on the effect of claim scope:} Our dataset consists of 8K book-level and 8K chapter-level claims. We fine-tune \prolongbase\ separately on each claim scope subset, resulting in \prolongftchapter\ and \prolongftbook.

\paragraph{Ablation on the effect of data length:}
Prior work shows that fine-tuning on short texts can improve long-context performance in tasks like QA and summarization \citep{dubey2024llama, gao2024trainlongcontextlanguagemodels}. Since our dataset contains long documents averaging 90K tokens, we test whether short-text fine-tuning also helps with long-context claim verification. We use WritingPrompts~\citep{fan-etal-2018-hierarchical}, a dataset of 300K stories averaging 742 tokens, and extract claims directly without generating outlines or summaries.\footnote{We use a prompt similar to the one in \S\ref{data:claims_generation}.} We collect 19K claims and train on \prolongbase\ to get \prolongwp.\footnote{After doing hyperparameter tuning on 2K training samples, we decide on the learning rate of 1e-5 and batch size of 16 as the best training configurations. We tested learning rates of 1e-5, 1e-6, 1e-7 and batch sizes of 8, 16, 32, 64.}


\subsection{Evaluation}

Beyond claim verification, we expect that training on our synthetic dataset will also improve performance on related tasks. Therefore, we  include both reasoning and narrative understanding benchmarks that vary in input lengths and tasks.

\paragraph{Claim verification:} To measure accuracy, we calculate the percentage of cases in which a model correctly verifies both true and false claims within a given pair.

\noindent \textbf{\includegraphics[height=0.6em]{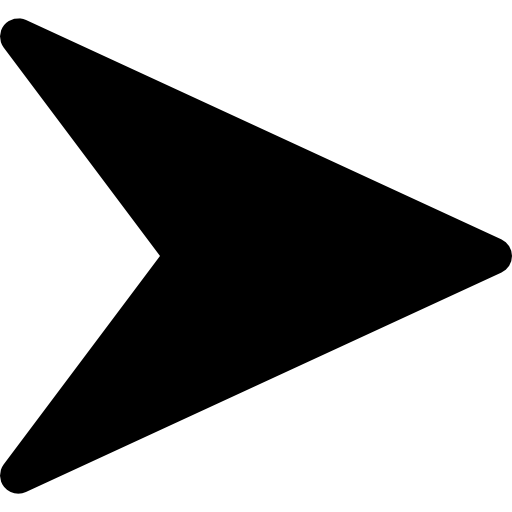} \pipeline-test} contains 1,000 true/false claim pairs drawn from 53 books, evenly split between book-level and chapter-level claims.

\noindent \textbf{\includegraphics[height=0.6em]{assets/graphics/right-arrow.png} NoCha} \citep{karpinska_one_2024} consists of 1,001 true/false claim pairs about recent fiction books (up to 336k tokens). These claims, crafted by annotators familiar with the books, are much harder to verify compared to those in \pipeline-test.


\paragraph{General narrative understanding:} We use three existing benchmarks as detailed below.

\noindent \textbf{\includegraphics[height=0.6em]{assets/graphics/right-arrow.png} NarrativeQA} \citep{kocisky_narrativeqa_2018} is a long-form Q\&A benchmark that requires models to process entire books or movie scripts to answer provided questions. The benchmark consists of 1,572 stories and summaries as well as 46,675 human-written questions. We use the HELMET implementation \citep{yen2024helmet} for this benchmark.

\noindent \textbf{\includegraphics[height=0.6em]{assets/graphics/right-arrow.png} $\infty$Bench QA} \citep{zhang2024inftybenchextendinglongcontext} is a long-form Q\&A benchmark that requires models to answer 351 questions about novels. We use the HELMET implementation but use GPT-4o's judgment as a metric instead of ROUGE F1 (see \S\ref{appendix:infbench-metric} for explanation).

\noindent \textbf{\includegraphics[height=0.6em]{assets/graphics/right-arrow.png} MuSR} \citep{sprague_musr_2024} includes 756 algorithmically generated problems such as murder mysteries, object placement questions, and team allocation optimization. We use the LM Harness~\citep{eval-harness} implementation.


%% file: sections/4-results.tex
\section{Results \& analysis}
Our fine-tuned models set a new state of the art for $<$10B models on long-context claim verification while also improving baseline performance on narrative understanding tasks. 

\subsection{\pipeline\ models outperform baselines on narrative claim verification} \label{subsec:main_results}

On \pipeline-test, our fine-tuned models significantly outperform the instruct models they are initialized from (referred to as baselines),\footnote{\prolongftbalanced\ is initialized from \prolongbase\ instead of \prolonginst. However, since performing evaluation intended for instruct models on a continually pretrained model may not be ideal, we exclude \prolongbase's results from Table \ref{tab:main-result}. As shown in Table \ref{tab:prolong-base-acc}, \prolongbase\ performs significantly worse than \prolonginst\ on \pipeline-test.} as shown in Table \ref{tab:main-result}. 

\begin{wrapfigure}{r}{0.5\linewidth} 
  \centering
  \vspace{-\baselineskip} 
  \includegraphics[width=\linewidth]{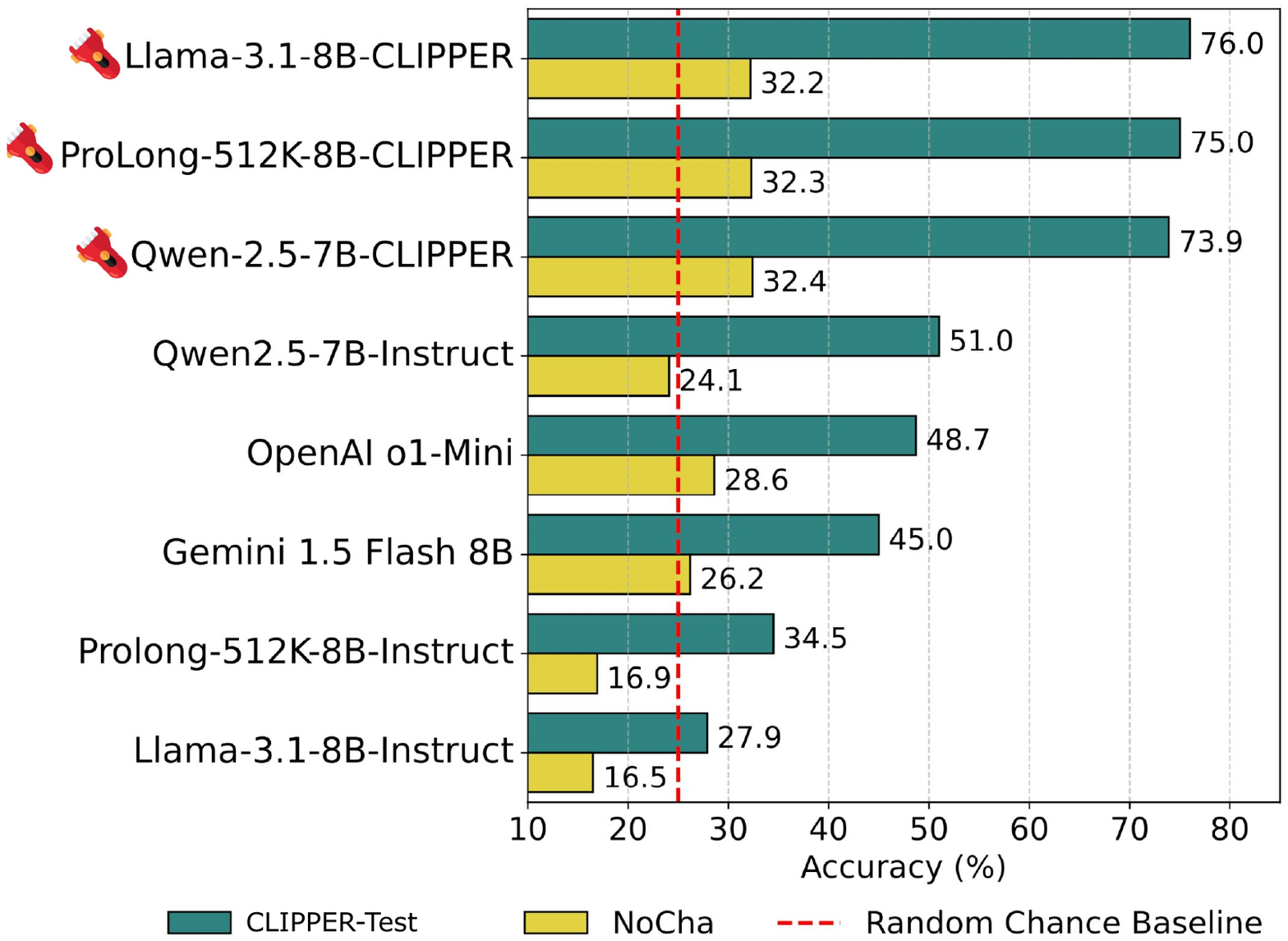}
  \caption{Results on \pipeline’s test set and NoCha for
baselines, small closed models, and \pipeline\ models.
Fine-tuning on our synthetic data significantly improves
narrative claim verification.}
\vspace{-\baselineskip}
\vspace{-\baselineskip}
\vspace{-\baselineskip}
  \label{fig:placeholder}
\end{wrapfigure}
For example, \qwenftbalanced\ achieves over a 20\% performance gain compared to \qweninst, while \llamaftbalanced\ sees nearly triple the performance of \llamainst. These substantial improvements demonstrate the effectiveness of \pipeline-generated data.







\begin{table*}[t]
\centering
\resizebox{\textwidth}{!}{%
\begin{tabular}{l c c c c c}
\toprule
\textbf{Models} & \textbf{\pipeline-test} & \textbf{NoCha} & \textbf{NarrativeQA} & \textbf{MuSR} & \boldmath$\infty$\textbf{Bench QA}\\
\midrule
Qwen2.5-7B-Instruct
  & 51.0\%  
  & 24.1\% 
  & 40.3\%  
  & 41.2\%
  & 35.3\%\\

Llama-3.1-8B-Instruct
  & 27.9\%  
  & 16.5\%
  & 47.7\%  
  & 40.3\%
  & \textbf{47.8\%}\\

ProLong-512K-8B-Instruct
  & 34.5\%  
  & 16.9\%
  & 44.0\%  
  & 42.3\%
  & 42.6\%\\
\midrule
\includegraphics[height=0.8em]{assets/graphics/elmo-clipper.png} Qwen2.5-7B-\pipeline\
  & 73.9\%  
  & \textbf{32.4\%}
  & 46.0\% 
  & \textbf{45.2\%}
  & 42.3\%\\

\includegraphics[height=0.8em]{assets/graphics/elmo-clipper.png} Llama-3.1-8B-\pipeline\ 
  & \textbf{76.0\%}
  & 32.2\%
  & \textbf{49.0\%}  
  & 43.6\%
  & 46.5\%\\

\includegraphics[height=0.8em]{assets/graphics/elmo-clipper.png} ProLong-512K-8B-\pipeline\
  & 75.0\% 
  & 32.3\%
  & \textbf{49.0\%}  
  & 44.5\%  
  & 38.5\%\\

ProLong-512K-8B-WritingPrompts
  & 63.0\%  
  & 24.1\%
  & 31.0\%  
  & \textbf{45.2\%}
  & 35.8\%\\
\bottomrule
\end{tabular}
}
\caption{Model accuracy on claim verification (\pipeline-test, NoCha) and narrative understanding benchmarks (NarrativeQA, MuSR, $\infty$Bench QA). Fine-tuning models using \pipeline\ improves performance on claim verification and narrative understanding. }
\label{tab:main-result}
\end{table*}

\paragraph{Fine-tuning on our data improves performance on NoCha:} A similar trend is observed on NoCha. The performance improvements range from an 8\% gain for strong baselines like \qweninst\ to a twofold increase for weaker baselines such as \llamainst\ and \prolonginst. It is worth noting that all baseline models initially perform below the random chance baseline of 25\%, but our fine-tuned models consistently surpass this threshold. 

\paragraph{Performance gap between \pipeline-test and NoCha:} We note that the performance gap between NoCha and \pipeline-test\ is likely due to the nature of the events involved in the claims. While \pipeline-test\ consists of synthetic claims derived from events in model-generated outlines, NoCha’s human-written claims may involve reasoning about low-level details that may not typically appear in such generated outlines. Future work could incorporate more low-level events into chapter outlines to create more diverse claims.
\subsection{Fine-tuning on \pipeline\ improves on other narrative reasoning tasks}  Beyond long-context reasoning, our models also show improvements in narrative understanding and short-context reasoning tasks. On NarrativeQA, which requires comprehension of movie scripts or full books, our best-performing models, \llamaftbalanced\ and \prolongftbalanced, achieve a 2\% and 5\% absolute improvement over their respective baselines. Similarly, on MuSR, a short-form reasoning benchmark, our strongest model, \qwenft, achieves 45.2\% accuracy, surpassing the 41.2\% baseline. However, on $\infty$Bench QA, only \qwenftbalanced\ outperforms the baseline by approximately 7\%. In contrast, \llamaftbalanced\ and \prolongftbalanced\ show slight performance declines of up to 4\%. Thus, while fine-tuning on \pipeline\ data improves performance on reasoning and some aspects of narrative understanding, its transferability is not universal across domains.

\subsection{Long-context claim data is more helpful than short-context data}
Our results stand in contrast to prior studies suggesting short-form data benefits long-context tasks \citep{dubey2024llama, gao2024trainlongcontextlanguagemodels} more than long data. While \prolongwp, trained on short data, outperforms baselines, it underperforms across all four long-context benchmarks compared to models fine-tuned on our data. This underscores the need for high-quality long-context data generation pipelines like \pipeline.

\subsection{Fine-tuning on CoTs results in more informative explanations}
We evaluate the groundedness of CoT reasoning generated by our fine-tuned models using DeepSeek-R1-Distill-Llama-70B (\S\ref{data:cot_validation}). Here, a reasoning chain is counted as grounded when every plot event in the chain can be found in the chapter outline that it cites. Table \ref{tab:cot-groundedness} shows that fine-tuning significantly improves groundedness across all models, with \prolongftbalanced\ achieving the highest rate (80.6\%), followed closely by \llamaftbalanced\ (75.9\%). Looking closer at the content of the explanations (Table \ref{tab:cot-examples}), the baseline model (\llamainst) often gives a generic response without citing any evidence, whereas \llamaftbalanced\ explicitly references Chapter 9 and specifies the cause-and-effect relationship.

\begin{table*}[t] 
    \centering 
    \scriptsize 
    \resizebox{\textwidth}{!}{
        \begin{tabular}{p{\textwidth}} 
        \toprule 
        \cellcolor{gray!10}\textbf{Claim:} "Jacques consistently uses pastoral themes to express his love for Belle-bouche, first through metaphorical discussions of shepherds and shepherdesses, and later while walking through an orchard and discussing Arcadian lovers." \\
        \midrule 
        \end{tabular}
    }
    \resizebox{\textwidth}{!}{
        \begin{tabular}[t]{p{0.45\textwidth} p{0.55\textwidth}} 
        \parbox[t]{0.45\textwidth}{%
        \textbf{\llamainst's CoT:} ``The statement claims that Jacques consistently uses pastoral themes to express his love for Belle-bouche. To evaluate this statement, we need to examine the text and identify instances where Jacques uses pastoral themes to express his love for Belle-bouche. {\color{purple}The text mentions} that Jacques discusses crooks, flowers, and shepherdesses with Belle-bouche, and later, while walking through an orchard, he talks about Arcadian lovers, specifically mentioning Strephon, Chloe, Corydon, Daphnis, and Narcissus. These instances suggest that Jacques does use pastoral themes to express his love for Belle-bouche. However, we need to consider whether this is a consistent pattern throughout the text." \newline \textit{Answer: TRUE}%
        } & 
        \parbox[t]{0.55\textwidth}{%
        \textbf{\llamaftbalanced's CoT:} ``Here are the relevant details from the text: \\ \vspace{-0.12in}
        \begin{itemize}[leftmargin=*]
            \item {\color{teal}Chapter 2}: Jacques engages in a metaphorical discussion about shepherds and shepherdesses, expressing his love for Belle-bouche.
            \item {\color{teal}Chapter 17}: Jacques and Belle-bouche walk through the orchard, discussing the twentieth of May as a day for lovers.
            \item Relationship: Recurring theme of pastoral love
        \end{itemize}
        In {\color{teal}Chapter 2}, Jacques engages in a metaphorical discussion about shepherds and shepherdesses to express his love for Belle-bouche. This theme continues in {\color{teal}Chapter 17}, where Jacques and Belle-bouche walk through an orchard, discussing the twentieth of May as a day for lovers. This connection shows Jacques' consistent use of pastoral imagery to convey his romantic feelings for Belle-bouche throughout the story." \newline \textit{Answer: TRUE}%
        } \\
        \bottomrule
        \end{tabular}
    }
    \caption{Comparison of CoT responses from \llamainst\ and \llamaftbalanced\ for a claim in John Esten Cooke's The Youth of Jefferson. \llamaftbalanced\ includes more explicit chapter references ({\color{teal}green}) and details than \llamainst\ ({\color{purple}red}).}
    \label{tab:cot-examples} 
\end{table*}

\begin{table*}[t]
\centering
\footnotesize
\resizebox{\textwidth}{!}{%
\begin{tabular}{p{0.1\textwidth}p{0.04\textwidth}p{0.43\textwidth}p{0.43\textwidth}}
\toprule
\multicolumn{1}{c}{\textsc{Category}} & \multicolumn{1}{c}{\textsc{Freq (\%)}} & \multicolumn{1}{c}{\textsc{True Claim}} & \multicolumn{1}{c}{\textsc{False Claim}} \\
\midrule
Event & 43.2 & The Polaris unit, initially assigned to test a new audio transmitter on Tara, explores the planet's surface {\color{teal}using a jet boat without landing}. & The Polaris unit, initially assigned to test a new audio transmitter on Tara, explores the planet's surface by {\color{purple}landing their spaceship}. \\
\midrule
Person & 31.6 & The cattle herd stolen from Yeager by masked rustlers is later found in {\color{teal}General Pasquale}'s possession at Noche Buena. & The cattle herd stolen from Yeager by masked rustlers is later found in {\color{purple}Harrison}'s possession at Noche Buena. \\
\midrule
Object & 15.8 & The alien structure Ross enters contains both a chamber with {\color{teal}a jelly-like bed} and {\color{teal}a control panel capable of communicating with other alien vessels}. & The alien structure Ross enters contains both a chamber with {\color{purple}a metal bed} and {\color{purple}a control panel capable of time travel}. \\
\midrule
Location & 13.7 & Costigan rescues Clio twice: first from Roger on his planetoid, and later from a {\color{teal}Nevian city} using a stolen space-speedster. & Costigan rescues Clio twice: first from Roger on his planetoid, and later from a {\color{purple}Triplanetary city} using a stolen space-speedster. \\
\midrule
Time & 6.3 & Jean Briggerland's meeting with ex-convicts Mr. Hoggins and Mr. Talmot, where she suggests a burglary target, {\color{teal}follows} a failed attempt on Lydia's life involving a speeding car on the sidewalk. & Jean Briggerland's meeting with ex-convicts Mr. Hoggins and Mr. Talmot, where she suggests a burglary target, {\color{purple}precedes} a failed attempt on Lydia's life involving a speeding car on the sidewalk. \\
\midrule
Affect & 4.2 & David Mullins, who initially expresses {\color{teal}skepticism} about Chester's hiring, later fires Chester on false pretenses and immediately replaces him with Felix. & David Mullins, who initially expresses {\color{purple}enthusiasm} about Chester's hiring, later fires Chester on false pretenses and immediately replaces him with Felix. \\
\bottomrule
\end{tabular}}
\caption{Taxonomy of perturbations causing false claims to be misclassified as true. True and false details are highlighted in {\color{teal}green} and {\color{purple}red}, respectively. Frequencies may exceed 100\% due to multi-labeling. See \S\ref{appendix:error-analysis} for definitions and analysis.}
\label{tab:error-analysis-dist}
\end{table*}

\subsection{Small models struggle with book-level reasoning} 
\label{subsection:chap-book-ft}
Trained only on 8K chapter-level claims, \prolongftchapter\ outperforms \prolongftbook\ on both chapter- and book-level test subsets (Table \ref{tab:chapter_vs_book}). This likely reflects the limitations of smaller models (7B/8B) in handling the complex reasoning required for book-level claims, in line with prior findings \citep{qi2024quantifyinggeneralizationcomplexitylarge}. The performance gap between the models is modest (4.2\%), and we leave exploration of larger models ($>$70B) to future work due to compute constraints.

\subsection{Fine-tuned models have a difficult time verifying False claims} \label{sec:error-analysis}
To study cases where fine-tuned models struggle, we analyze \qwenftbalanced\ outputs. Among 1,000 book-level claim pairs in \pipeline-test, the model fails to verify 37 true claims and 97 false claims, aligning with NoCha findings \citep{karpinska_one_2024} that models struggle more with false claims. We investigate perturbations that make false claims appear true and present a taxonomy with examples in Table \ref{tab:error-analysis-dist}, with further details in \S\ref{appendix:error-analysis}.

%% file: sections/5-related.tex
\section{Related work}  
\paragraph{Long-context language modeling:}
The context size of LLMs has expanded significantly~\citep{openai2024gpt4technicalreport, dubey2024llama, geminiteam2024gemini15unlockingmultimodal, yang_qwen25-1m_2025}, thanks to position inter- and extrapolation techniques~\citep{press_train_2022, su_roformer_2023, peng_yarn_2023} and efficient attention implementation~\citep{dao2022flashattentionfastmemoryefficientexact, dao2023flashattention2fasterattentionbetter,liu2023blockwise}. Longer data has been used during continual pretraining~\citep{lieber2024jambahybridtransformermambalanguage, xiong_effective_2023} or alignment stage \citep{bai_longalign_2024, xiong2024artificialneedlesrealhaystacks,an2024makellmfullyutilize}. \pipeline\ augments existing long-form book texts with synthetic but challenging claims, which serves as the foundation of a fine-tuning pipeline to improve LLMs’ understanding and reasoning over long-context data.


\paragraph{Instruction-tuning data generation:} 
Short-form data generation methods either induce instruction data from texts~\citep{honovich2022instructioninductionexamplesnatural, zhou2023largelanguagemodelshumanlevel, li2024selfalignmentinstructionbacktranslation} or generate instruction-output pairs simultaneously~\citep{wang2023selfinstructaligninglanguagemodels}. Long-context data is synthesized through induction from long-form documents~\citep{pham_suri_2024, koksal_longform_2023}, random document segments~\citep{xiong_effective_2023}, or bootstrapping short documents~\citep{an2024makellmfullyutilize, xu2024chatqa2bridginggap, wu2024longcontextalignmentshort, wang2024bootstrapcontextlength}. \pipeline\ uses instruction induction from compressed document representations to create instruction-tuning data for long-context LLMs.

\paragraph{Reasoning alignment:}
Previous work includes inference-time scaling~\citep{openai_o1_2024, deepseekai2025deepseekr1incentivizingreasoningcapability, muennighoff2025s1simpletesttimescaling}, prompting~\citep{wei_chain--thought_2023, kojima2023largelanguagemodelszeroshot,yao2023treethoughtsdeliberateproblem, wang2023selfconsistencyimproveschainthought}, and fine-tuning LLMs on reasoning data~\citep{chung_scaling_2022, huang-etal-2023-large, puerto2024finetuningdivergentchainsthought, yeo2025demystifyinglongchainofthoughtreasoning}. These reasoning data are either human-written rationale~\citep{alkhamissi_opt-r_2023} or chain of thoughts distilled from larger models~\citep{hsieh-etal-2023-distilling,li-etal-2023-symbolic, ho-etal-2023-large, zelikman2022starbootstrappingreasoningreasoning}. We find that fine-tuning models on CoTs improves the generated explanations for the validity of book claims.

%% file: sections/6-conclusion.tex
\section{Conclusion}
We introduce \pipeline, a compression-based pipeline for generating synthetic narrative claims. We create 19K true/false claims at both book and chapter levels. Our fine-tuned models set a new state-of-the-art among $<$10B models on claim verification and achieve improvement on narrative understanding tasks. Future work could examine the effect of book-level claims on larger models and explore methods for generating harder claims to better match human-written benchmarks like NoCha.

%% file: sections/7-limitations-ethics.tex
\section*{Limitations}

We only perform hyperparameter tuning on ProLong-Base due to the high cost of the training process. To put things into perspective, training a model on our full test set requires approximately 50 hours on 8 A100 GPUs, each costing \$2 per hour to rent. Even training on our tuning subset takes 6 hours. Therefore, extending training further is prohibitively expensive.

Similarly, we do not hire human annotators to write claims for our dataset due to the prohibitive cost and the need for numerous annotators who have thoroughly read the books (\autoref{tab:cost_analysis}). While this decision may result in less complex claims, our approach offers greater adaptability to new books while significantly reducing costs.

The compression stage can be challenging to fine-tune, subject to model biases, and prone to potential hallucinations. Due to the large volume of data, verifying its accuracy is also difficult, whether via prompting or human annotation. However, we note that prior research has demonstrated that LLMs are capable of producing high-quality summaries of long documents \cite{chang_booookscore_2024, kim_fables_2024}. In addition, these compressed representations could still provide a strong foundation for claim generations, as most of \pipeline's claims are grounded in the original book (\autoref{data:claims_generation}). 

%% file: sections/A-data.tex
\newpage
\appendix
\section{Data Collection}

\subsection{Books used in manual analysis}
Table \ref{tab:six-books} lists six books used in our manual analysis. These books are chosen due to the annotator's familiarity with the content, which eases the manual verification process. 
\begin{table*}[htbp]
    \small
    \centering
    \resizebox{\textwidth}{!}{%
    \begin{tabular}{llccc}
        \toprule
        Title & Author & Publication Year & Number of Tokens & Number of Chapters \\
        \midrule
        Anne of the Island & L. M. Montgomery & 1915 & 111,337& 41 \\
        Alice in Wonderland & Lewis Carroll & 1865 & 36,691 & 12 \\
        The Murder of Roger Ackroyd & Agatha Christie & 1926 & 98,602 & 27 \\
        The Picture of Dorian Gray & Oscar Wilde & 1890 & 105368 & 20\\
        Frankenstein & Mary Shelley & 1818 & 97,574 & 24\\
        The Adventures of Tom Sawyer & Mark Twain & 1876 & 97,968 & 35 \\
        \bottomrule
    \end{tabular}
    }
    \caption{Six books used in our manual analysis. Books are chosen due to familiarity with the content.}
    \label{tab:six-books}
\end{table*}

\subsection{Does memorization have an effect on claim verification performance?}
\label{appendix:memorization}
We measure the performance of the models used for fine-tuning on our test set, with and without book text. We provide the book title and author name where the book text is not provided. Our hypothesis is that if the model does better than the random chance baseline (25\% accuracy) without the book text, then the claims are either too easy or can be verified without even reasoning over the texts. 
\begin{table}[htbp]
\small
    \centering
    \begin{tabular}{lcc}
        \toprule
        Models & No Text & With Text\\
        \midrule
        \prolonginst\ & 0.0\%& 35.6\%\\
        \llamainst\ & 0.0\%& 32.8\%\\
        \qweninst\ & 0.0\%& 51.4\%\\
        \bottomrule
    \end{tabular}
    \caption{Accuracy on \pipeline's test set (with and without book texts).}
    \label{tab:memorization}
\end{table}

Table \ref{tab:memorization} shows that all baseline models perform below random chance, significantly trailing behind the performance achieved when the book text is included in the claim verification prompt. These results indicate that even if a model has memorized the book texts or generated claims, such memorization does not affect its performance on the task itself.

\subsection{Are the True/False claims distinguishable without the book texts?}
We ask the question of whether distinguishing between True and False claims is inherently too easy. If so, then the high performance of the fine-tuned models may be attributed merely to their ability to detect formatting cues rather than actually reasoning. To investigate this, we prompt both baseline and fine-tuned models to verify claims without providing any book texts or metadata. Our hypothesis is that if a model performs better than random chance under these conditions, then the claims are likely too easily distinguishable based on their formatting alone. 

\begin{table}[htbp]
    \small
    \centering
    \begin{tabular}{lcc}
        \toprule
        Models & Before SFT & After SFT\\
        \midrule
        \prolonginst\ &  0.0\%& 25.2\%\\
        \llamainst\ & 20.2\%& 13.8\%\\
        \qweninst\ & 21.7\%& 22.9\%\\
        \bottomrule
    \end{tabular}
    \caption{Accuracy on \pipeline's test set (no book text or metadata provided).}
    \label{tab:claim-formatting}
\end{table}

As shown in Table \ref{tab:claim-formatting}, even after fine-tuning, the models perform only marginally above random guessing. We conclude that, without the contextual information from the book text, True/False claims are not easily distinguishable.

\subsection{Cost Analysis}
\label{appendix:data-cost}
Table \ref{tab:cost_analysis} shows the cost incurred by running each stage of our data synthesis pipeline. With the exception of deduplication, which is done by GPT-4o, each stage of the pipeline is performed by Claude.  Table \ref{tab:cost-naive-main} shows the estimated per claim cost for the \texttt{na\"ive} versus \texttt{main} approach based on estimated cost for 6 books. For human annotation, NoCha \citep{karpinska_one_2024} reports that their total cost of annotating 1,001 claim pairs is \$3,327 USD, so each claim costs around \$1.7.

\begin{table}[htbp]
    \centering
    \small
    \begin{tabular}{lc}
        \toprule
        Stage & Cost\\
        \midrule 
        Book summary generation & \$0.0021 \\
        Chapter outline generation & \$0.0107 \\
        Book-level claim synthesis & \$0.0129 \\
        Chapter-level claim synthesis & \$0.0172 \\
        Deduplication & \$0.0021 \\
        Verification & \$0.0064 \\
        \midrule
        Total&\$0.0514\\
        \bottomrule\\
    \end{tabular}
    \caption{Cost to run pipeline per claim (in US dollars, rounded to four decimal places).}
    \label{tab:cost_analysis}
\end{table}

\begin{table}[htbp]
\centering
\small
\begin{tabular}{@{}lcc@{}}
\toprule
                            & \naive & \pipeline \\ \midrule
Cost per claim (book-level) & \$0.09         & \$0.07        \\
Cost per claim (chap-level) & \$0.04         & \$0.02        \\
\bottomrule
\end{tabular}
\caption{Estimated cost for our \naive\ vs \pipeline\ approach (rounded to two decimal places)}
\label{tab:cost-naive-main}
\end{table}

\subsection{Prompts}
\label{appendix:data-construction}
Table \ref{tab:prompt-mapping} shows stages to construct \pipeline, mapped to their corrresponding prompts.
\begin{table}[htbp]
    \small
    \centering
    \begin{tabular}{ll}
        \toprule
         Prompt & Figure\\
         \midrule
         Chapter outline generation & \ref{fig:chapter-outline-prompt}\\
         Book summary generation & \ref{fig:summary-prompt} \\
         Chapter-level claim extraction & \ref{fig:chapter-level} \\
         Book-level claim extraction & \ref{fig:book-level} \\
         Claim deduplication & \ref{fig:dedup-prompt} \\
         Claim verification & \ref{fig:verification-prompt-1}, \ref{fig:verification-prompt-2}, \ref{fig:verification-prompt-3}, \ref{fig:verification-prompt-4}, \ref{fig:verification-prompt-5}\\
         \midrule
         Chapter-level claim extraction (\naive) & \ref{fig:chapter-alt}\\
         Book-level claim extraction (\naive) & \ref{fig:book-alt}\\
         \bottomrule
    \end{tabular} 
    \caption{Figure references for each prompt.}
    \label{tab:prompt-mapping}
\end{table}

\begin{figure*}[htbp]
\centering
\begin{tcolorbox}[colback=gray!5!white, colframe=black, title=Prompt for Chapter Outline Generation]
\lstset{
    basicstyle=\ttfamily\footnotesize,
    breaklines=true,
    frame=none,
    xleftmargin=0pt,
    framexleftmargin=0pt,
    columns=fullflexible,
    tabsize=1,
    breakindent=0pt,
    breakautoindent=false,
    postbreak=\space,
    showstringspaces=false,
}
\begin{lstlisting}[language=Markdown]
Your task is to create a detailed and objective outline for Chapter {order} of a book. You will be provided with the text of Chapter {order}. Ensure that your outline faithfully represents the content of the text.

First, carefully read the text of Chapter {order}:
<current_chapter>
{curr}
<current_chapter>

Finally, create an outline for Chapter {order} using the following format:

<synopsis>A one-sentence summary of the current chapter.</synopsis>
<events>A chronological list of at most 7 major events in the chapter. The list should formatted as a numbered list. Each event should be one sentence long, describing specific details on what happens, where it happens, and which characters are involved. DO NOT include subjective interpretation of the events.</events>
<characters>A numbered list of characters in the chapter. Include only those that are mentioned in the major events. Each character should have the format: [character full name]: [character role/relationship with the main characters], [physical appearance (if mentioned)], [personality (if mentioned)], first seen at [the first setting the character is in], last seen at [last setting the character is in].</characters>

Now, create an objective outline for Chapter {order} based on the provided text. Ensure that your outline is concise, coherent, and accurately represents the content of the chapter.

<synopsis>
[One-sentence summary of the current chapter.]
</synopsis>
<events>
1. [Event 1: Specific details on specific details on what happens, where it happens, and which characters are involved.]
2. [Event 2: Specific details on what happens, where it happens, and which characters are involved.]
3. [Event 3: Specific details on what happens, where it happens, and which characters are involved.]
4. [Event 4: Specific details on what happens, where it happens, and which characters are involved.]
5. [Event 5: Specific details on what happens, where it happens, and which characters are involved.]
6. [Event 6: Specific details on what happens, where it happens, and which characters are involved.]
7. [Event 7: Specific details on what happens, where it happens, and which characters are involved.]
</events>
<characters>
1. [Character 1: Character role/relationship with the main characters, physical appearance (if mentioned), personality (if mentioned), first seen at the first setting the character is in, last seen at the last setting the character is in.]
(Repeat the format for each character mentioned in the major events.)
</characters>

Remember to focus on the objective representation of the chapter content and avoid adding personal opinions or interpretations. Good luck!
\end{lstlisting}
\end{tcolorbox}
\caption{Prompt for generating chapter outlines in our dataset.}
\label{fig:chapter-outline-prompt}
\end{figure*}

\begin{figure*}[htbp]
\centering
\begin{tcolorbox}[colback=gray!5!white, colframe=black, title=Prompt for Generating Book Summary]
\lstset{
    basicstyle=\ttfamily\footnotesize,
    breaklines=true,
    frame=none,
    xleftmargin=0pt,
    framexleftmargin=0pt,
    columns=fullflexible,
    tabsize=1,
    breakindent=0pt,
    breakautoindent=false,
    postbreak=\space,
    showstringspaces=false,
}
\begin{lstlisting}[language=Markdown]
Your task is to write a summary for the book below, make sure to include vital information related to key events, backgrounds, settings, characters, their objectives, and motivations. You must briefly introduce characters (with their full name), places, and other major elements if they are being mentioned for the first time in the summary. The book may feature non-linear narratives, flashbacks, switches between alternate worlds or viewpoints, etc. Therefore, you should organize the summary so it presents a consistent and chronological narrative. The summary must be within 1000 words. The summary should span multiple paragraphs and should be written as a single continuous narrative, not as a list of bullet points or an outline. DO NOT include the book name in the summary.

#Book 
{book}

#Summary
\end{lstlisting}
\end{tcolorbox}
\caption{Prompt for generating chapter outlines in our dataset.}
\label{fig:summary-prompt}
\end{figure*}

\begin{figure*}[htbp]
\centering
\begin{tcolorbox}[colback=gray!5!white, colframe=black, title=Prompt for Extracting Chapter-level Claims]
\lstset{
    basicstyle=\ttfamily\scriptsize,
    breaklines=true,
    frame=none,
    xleftmargin=0pt,
    framexleftmargin=0pt,
    columns=fullflexible,
    tabsize=1,
    breakindent=0pt,
    breakautoindent=false,
    postbreak=\space,
    showstringspaces=false,
}
\begin{lstlisting}[language=Markdown]
Your task is to create factual statements that incorporate multiple events from Chapter {chapter} of a book. These factual statements must be objective and specific, grounded in the specific chapter, and consistent with the entire book. The included events must be specific. The facts cannot contain any interpretations, speculations, or subjective statements. In addition to each fact, provide a minimally corrupted version of the fact, which sounds plausible but is wrong based on the entire book.
To create valid facts, follow these guidelines step-by-step:
1. Carefully read through the book.
2. In a <brainstorm> section: 
    - Identify different events that are strongly related to one another in a single chapter.
    - Consider the relationship between these events, but DO NOT include this brainstorming in the resulting fact. If a meaningful relationship is found, move on to the next step; otherwise, proceed to the next fact.
3. Formulate your facts based on this analysis. Ensure that your facts:
    - Contain a single sentence
    - Are coherent with the entire book
    - Include multiple detailed and strongly related events from a single chapter in the book
    - Are self-contained and independent of other facts
    - No part of the fact should be subjective interpretations or speculations
    - Do not contradict or duplicate existing facts
    - Do not contain chapter information (e.g. "In Chapter x, ...")
    - Do not contain quotes from the book. 
4. Formulate a minimally corrupted version of each fact. 
    - Only one aspect of the fact, such as an atomic detail or the relationship between details, should be altered.
    - The corrupted fact should sound plausible but be clearly wrong based on the entire book.
    - The corrupted fact should be similar with the original fact in terms of length and sentence structure to make it harder to verify solely based on surface-level features. 
First, read the book: 
<book>{book}</book>
Finally, review the existing facts:
<existing_facts>{existing_facts}</existing_facts>
Now, generate as many valid facts as possible based on the provided chapter. Return "No meaningful fact" if there is no valid fact. Present your facts in the following format:
<facts>
<fact_1>
<brainstorm>[Your brainstorm notes here]</brainstorm>
Fact: [Your fact here]
Fact Reasoning: [Your explanation here (including the chapter associated with each event)]
Source: [Chapter involved]
Corrupted Fact: [Your corrupted fact here]
Corrupted Fact Reasoning: [Your explanation here]
</fact_1>
[Continue with additional facts...]
</facts>
Remember to create facts that are objectively valid, coherent with the entire book, and demonstrate strong, meaningful relationships between the events from Chapter {chapter}. No part of the fact can include subjective interpretations or generalizations. The relationship must be meaningful. The included events must be SPECIFIC and DETAILED!!!!!
\end{lstlisting}
\end{tcolorbox}
\caption{Prompt for extracting chapter-level claims}
\label{fig:chapter-level}
\end{figure*}

\begin{figure*}[htbp]
\centering
\begin{tcolorbox}[colback=gray!5!white, colframe=black, title=Prompt for Extracting Book-level Claims]
\lstset{
    basicstyle=\ttfamily\footnotesize,
    breaklines=true,
    frame=none,
    xleftmargin=0pt,
    framexleftmargin=0pt,
    columns=fullflexible,
    tabsize=1,
    breakindent=0pt,
    breakautoindent=false,
    postbreak=\space,
    showstringspaces=false,
}
\begin{lstlisting}[language=Markdown]
Your task is to create factual statements that incorporate multiple events from multiple chapters of a book. These factual statements must be objective and specific, grounded in the involved chapters, and consistent with the entire book. The included events must be specific. The facts cannot contain any interpretations, speculations, or subjective statements. In addition to each fact, provide a minimally corrupted version of the fact, which sounds plausible but is wrong based on the entire book.
To create valid facts, follow these guidelines step-by-step:
1. Carefully read through the book.
2. In a <brainstorm> section: 
    - Identify different events that are strongly related to one another in multiple chapters.
    - Consider the relationship between these events, but DO NOT include this brainstorming in the resulting fact. If a meaningful relationship is found, move on to the next step; otherwise, proceed to the next fact.
3. Formulate your facts based on this analysis. Ensure that your facts:
    - Contain a single sentence
    - Are coherent with the entire book
    - Include multiple detailed and strongly related events from multiple chapters in the book
    - Are self-contained and independent of other facts
    - No part of the fact should be subjective interpretations or speculations
    - Do not contradict or duplicate existing facts
    - Do not contain chapter information (e.g. "In Chapter x, ...")
    - Do not contain quotes from the book. 
4. Formulate a minimally corrupted version of each fact. 
    - Only one aspect of the fact, such as an atomic detail or the relationship between details, should be altered.
    - The corrupted fact should sound plausible but be clearly wrong based on the entire book.
    - The corrupted fact should be similar with the original fact in terms of length and sentence structure to make it harder to verify solely based on surface-level features. 
First, read the book: 
<book>{book}</book>
Finally, review the existing facts:
<existing_facts>{existing_facts}</existing_facts>
Now, generate as many valid facts as possible based on the provided book. Return "No meaningful fact" if there is no valid fact. Present your facts in the following format:
<facts>
<fact_1>
<brainstorm>[Your brainstorm notes here]</brainstorm>
Fact: [Your fact here]
Fact Reasoning: [Your explanation here (including the chapter associated with each event)]
Source: [Chapters involved]
Corrupted Fact: [Your corrupted fact here]
Corrupted Fact Reasoning: [Your explanation here]
</fact_1>
[Continue with additional facts...]
</facts>
Remember to create facts that are objectively valid, coherent with the book, and demonstrate strong, meaningful relationships between the events from multiple chapters. No part of the fact can include subjective interpretations or generalizations. The relationship must be meaningful. The included events must be SPECIFIC and DETAILED!!!!!
\end{lstlisting}
\end{tcolorbox}
\caption{Prompt for extracting book-level claims}
\label{fig:book-level}
\end{figure*}

\begin{figure*}[htbp]
\centering
\begin{tcolorbox}[colback=gray!5!white, colframe=black, title=Prompt for Deduplicating Claims]
\lstset{
    basicstyle=\ttfamily\footnotesize,
    breaklines=true,
    frame=none,
    xleftmargin=0pt,
    framexleftmargin=0pt,
    columns=fullflexible,
    tabsize=1,
    breakindent=0pt,
    breakautoindent=false,
    postbreak=\space,
    showstringspaces=false,
}
\begin{lstlisting}[language=Markdown]
You will be given a list of facts. Your task is to identify all duplicate facts within this list. A fact is considered a duplicate if it is exactly the same as another fact, or if it contains the same information as another fact, even if worded differently.

Here is the list of facts:

<fact_list>
{fact_list}
</fact_list>

To identify the duplicate facts, follow these guidelines step by step:
1. Read through the list of facts carefully.
2. Identify any facts that are exact duplicates of each other, or that convey the same atomic information using different wording.
3. Return a list of facts that are duplicates of one other, along with an explanation of why they are duplicates.
4. DO NOT return facts that are not duplicates of any other fact in the list.

Here's an example of how your output should look:

<example>
<example_fact_list>
1. Jim worked hard, so he got a promotion.
2. Jim's hard work paid off, and he was promoted.
3. Jim and Sarah worked together on the project.
4. Jim worked hard, so he received praise from his boss, who promoted him.
</example_fact_list>

<example_answer>
- 1, 2: These two facts convey the same atomic information but are worded differently. 
</example_answer>
</example>

Remember, your goal is to identify all duplicate facts, whether they are exact matches or convey the same information in different words. Be thorough in your analysis and clear in your explanations. If there is no duplication in the list, output "No duplicates found."

<answer>
- [Index of duplicate facts, separated by commas]: [Explanation of why they are duplicates]
- [Index of duplicate facts, separated by commas]: [Explanation of why they are duplicates]
(... and so on)
</answer>
\end{lstlisting}
\end{tcolorbox}
\caption{Prompt for de-duplicating claims}
\label{fig:dedup-prompt}
\end{figure*}

\begin{figure*}[htbp]
\centering
\begin{tcolorbox}[colback=gray!5!white, colframe=black, title=Prompt for Verifying Claims with GPT-4o (Part 1)]
\lstset{
    basicstyle=\ttfamily\footnotesize,
    breaklines=true,
    frame=none,
    xleftmargin=0pt,
    framexleftmargin=0pt,
    columns=fullflexible,
    tabsize=1,
    breakindent=0pt,
    breakautoindent=false,
    postbreak=\space,
    showstringspaces=false,
}
\begin{lstlisting}[firstline=1, lastline=30, language=Markdown]
You will receive a book summary, a chapter outline, and a claim extracted from that outline. Your task is to verify whether the claim contains detailed information, presents a meaningful relationship, and shows consistency with both the book summary and chapter outline.

To verify the claim, follow these steps:
1. Read the summary, outline, and claim carefully to understand the context.
2. Decompose the claim into atomic parts. 
3. Analyze each atomic part:
    a. Is the part grounded in the events? 
    b. Does this part contradict any information in the summary or outline? Keep in mind that some books may have discontinuous plots or events, so just because a detail is mentioned before or after another in the summary does not mean they are temporally related.
4. Evaluate the relationship between the atomic parts:
    a. Is the relationship objectively valid and meaningful?
    b. Is the relationship a subjective interpretation or assumption not explicitly stated in the summary or outline?
    c. Does the relationship make sense based on the book summary and chapter outline?
5. Based on your analysis, provide your reasoning and verification result. Your reasoning should explain why you believe the claim is or is not valid based on the information provided in the book summary and chapter outline.

First, read the book summary:
<book_summary>
{book_summary}
</book_summary>

Next, review the chapter outline:
<chapter_outline>
{chapter_outline}
</chapter_outline>

Finally, consider the following claim:
<claim>
{claim}
</claim>

Here are two examples of valid and invalid claims:
\end{lstlisting} 
\end{tcolorbox}
\caption{Prompt for verifying claims with GPT-4o (Part 1)}
\label{fig:verification-prompt-1}
\end{figure*}

\begin{figure*}[htbp]
\centering
\begin{tcolorbox}[colback=gray!5!white, colframe=black, title=Prompt for Verifying Claims with GPT-4o (Part 2)]
\lstset{
    basicstyle=\ttfamily\footnotesize,
    breaklines=true,
    frame=none,
    xleftmargin=0pt,
    framexleftmargin=0pt,
    columns=fullflexible,
    tabsize=1,
    breakindent=0pt,
    breakautoindent=false,
    postbreak=\space,
    showstringspaces=false,
}
\begin{lstlisting}[firstline=31, lastline=60, language=Markdown]
<example_1>
<example_summary>
Laura Hand, Daniel Knowe, and Mo Gorch mysteriously return from a realm of death with altered memories, orchestrated by their enigmatic music teacher, Mr. Anabin, and the sinister Bogomil. As they grapple with their new realities, including Mo's discovery of his grandmother's death and Daniel's complex feelings for Laura's sister, Susannah, they face trials set by Anabin and Bogomil to remain in the living world. Alongside Bowie, another returned soul, they uncover the truth about their deaths while dealing with eerie supernatural encounters. Laura's newfound magical abilities strain her relationship with Susannah, leading to increasing tensions as Susannah begins to remember the truth.
As the trio navigates their altered lives, they become entangled in a larger, dangerous game orchestrated by Malo Mogge, Anabin, and Bogomil, who guard the door between life and death. Mo and Susannah discover that a Harmony guitar, hidden by Susannah, is the key sought by Malo Mogge, a powerful entity seeking immense power. The story culminates in a chaotic battle in Lovesend, where Laura, consumed by grief, vows to kill Malo Mogge. After absorbing Mogge's magic and becoming a powerful goddess, Laura faces the challenge of balancing her divine powers with her passion for music, while the other characters embrace their new roles. 
</example_summary>

<example_chapter_outline>
## Chapter 15 outline
1. Susannah gets frustrated about Laura and Daniel being close to each other and smashes Laura's old guitar in a fit of anger and frustration.
2. Mo eats a breakfast casserole made by Jenny and buys doughnuts and bagels on his way to Laura's house, feeling a mix of hunger and sadness.
3. Mo arrives at the Hands' house, where Laura and Daniel are eating ramen to satisfy their unusual hunger. 
4. Daniel reveals that he and Laura have swapped ears due to Mr. Anabin's magic, and they discuss the implications of this mistake.
5. Mo shares his encounter with a mysterious figure outside his house, leading to a heated discussion about Bogomil and Mr. Anabin.
6. The trio creates a list of goals to navigate their situation, including staying alive, figuring out how they died, and learning to do magic.
7. They attempt to perform magic by trying to transform a saltshaker into a hairless cat but fail, leading to further frustration.
8. Mo leaves the Hands' house, and they discover that the entire yard and house are covered in thousands of moths, adding to the surreal nature of their situation.
9. Laura finds the broken guitar pieces in her room, causing confusion and suspicion among the trio.
</example_chapter_outline>

<example_claim_1>
Laura finds the broken guitar pieces in her room, which Susannah smashed in a fit of anger and frustration.
</example_claim_1>

<example_verification_1>
<reasoning>
1. Analysis of each atomic part: 
    a. Laura finds the broken guitar pieces in her room: This part is grounded in event 9 of Chapter 15. 
    b. Susannah smashed the guitar in a fit of anger and frustration: This part is grounded in event 1 of Chapter 15.
2. Analysis of the relationship between atomic parts:
    a. The relationship is temporal, and thus valid. 
\end{lstlisting} 
\end{tcolorbox}
\caption{Prompt for verifying claims with GPT-4o (Part 2)}
\label{fig:verification-prompt-2}
\end{figure*}

\begin{figure*}[htbp]
\centering
\begin{tcolorbox}[colback=gray!5!white, colframe=black, title=Prompt for Verifying Claims with GPT-4o (Part 3)]
\lstset{
    basicstyle=\ttfamily\footnotesize,
    breaklines=true,
    frame=none,
    xleftmargin=0pt,
    framexleftmargin=0pt,
    columns=fullflexible,
    tabsize=1,
    breakindent=0pt,
    breakautoindent=false,
    postbreak=\space,
    showstringspaces=false,
}
\begin{lstlisting}[firstline=61, lastline=90, language=Markdown]
    b. The relationship is explicit and not a subjective interpretation.
    c. The relationship is grounded in the chapter outline. There is no contradicting information in the summary. 
Since all considerations are satisfied, the claim is VALID.
</reasoning>
<result>
VALID
</result>
</example_verification_1>

<example_claim_2>
Mo shares his encounter with a mysterious figure because Laura and Daniel discuss the implications of their swapped hands due to Mr. Anabin's magic.
</example_claim_2>

<example_verification_2>
<reasoning>
1. Analysis of atomic parts:
    a. Mo shares his encounter with a mysterious figure: This part is grounded in event 5 of Chapter 15.
    b. Laura and Daniel discuss the implications of their swapped hands due to Mr. Anabin's magic: There is no mention of hands being swapped. Even though event 4 discusses the swapping of ears, it does not relate to hands.
2. Analysis of the relationship between atomic parts:
    a. The relationship is NOT VALID because there is no direct link between Mo sharing his encounter and Laura and Daniel discussing the implications of their swapped hands. 
    b. The relationship is a subjective interpretation and not explicitly grounded in the summary or outline.
    c. The relationship is grounded in the chapter outline. There is no contradicting information in the summary.
Since 1a., 2a., and 2b. are not satisfied, the claim is INVALID.
</reasoning>
<result>
INVALID
</result>
</example_verification_2>

\end{lstlisting} 
\end{tcolorbox}
\caption{Prompt for verifying claims with GPT-4o (Part 3)}
\label{fig:verification-prompt-3}
\end{figure*}

\begin{figure*}[htbp]
\centering
\begin{tcolorbox}[colback=gray!5!white, colframe=black, title=Prompt for Verifying Claims with GPT-4o (Part 4)]
\lstset{
    basicstyle=\ttfamily\footnotesize,
    breaklines=true,
    frame=none,
    xleftmargin=0pt,
    framexleftmargin=0pt,
    columns=fullflexible,
    tabsize=1,
    breakindent=0pt,
    breakautoindent=false,
    postbreak=\space,
    showstringspaces=false,
}
\begin{lstlisting}[firstline=91, lastline=120, language=Markdown]
<example_claim_3>
Jenny cooks breakfast for Mo because he feels a mix of hunger and sadness.
</example_claim_3>

<example_verification_3>
<reasoning>
1. Analysis of atomic parts:
    a. Jenny cooks breakfast for Mo: This part is grounded in event 2 of Chapter 15.
    b. Mo feels a mix of hunger and sadness: This part is grounded in event 2 of Chapter 15.
2. Analysis of the relationship between atomic parts:
    a. The relationship is INVALID. There is no indication that Jenny cooked breakfast for Mo because he felt a mix of hunger and sadness. The events are happening simultaneously but are not causally connected.
    b. The relationship is a subjective interpretation and not explicitly grounded in the summary or outline.
    c. The relationship is grounded in the chapter outline. There is no contradicting information in the summary.
Since 2a. and 2b. are not satisfied, the claim is INVALID.
</reasoning>
<result>
INVALID
</result>   
</example_verification_3>
</example_1>

<example_2>
<example_summary>
Sarah Lee discovers an ancient wooden box with strange symbols in her attic, which contains a journal revealing the history of a secret society and a prophecy about the return of a powerful being known as "The Shadow." As unusual events begin to plague her town, Sarah, along with her friend Mark, uncovers clues that connect these occurrences to the prophecy. They find a hidden chamber beneath the town containing ancient texts and artifacts, including a weapon capable of banishing "The Shadow." With this weapon, they confront a member of the secret society who attempts to summon "The Shadow," and after a tense battle, Sarah successfully uses the weapon to banish the being, restoring peace to the town.

Throughout the story, the connection between the ancient artifact, the journal, and the unfolding events reveals the central role of the wooden box and the secret society in the impending danger. Sarah and Mark's journey highlights their struggle to protect their town from supernatural forces while deciphering the mysterious symbols and prophecies tied to the powerful entity, "The Shadow."
</example_summary>

<example_chapter_outline>
## Chapter 3 outline
\end{lstlisting} 
\end{tcolorbox}
\caption{Prompt for verifying claims with GPT-4o (Part 4)}
\label{fig:verification-prompt-4}
\end{figure*}

\begin{figure*}[htbp]
\centering
\begin{tcolorbox}[colback=gray!5!white, colframe=black, title=Prompt for Verifying Claims with GPT-4o (Part 5)]
\lstset{
    basicstyle=\ttfamily\footnotesize,
    breaklines=true,
    frame=none,
    xleftmargin=0pt,
    framexleftmargin=0pt,
    columns=fullflexible,
    tabsize=1,
    breakindent=0pt,
    breakautoindent=false,
    postbreak=\space,
    showstringspaces=false,
}
\begin{lstlisting}[firstline=121, lastline=160, language=Markdown]
1. Sarah discovers an ancient artifact in her attic, an intricately carved wooden box with strange symbols.
2. She finds an old journal in the box, detailing the history of a secret society that once protected the town.
3. The journal reveals a prophecy about the return of a powerful being known as "The Shadow."
4. Sarah decides to keep the discovery to herself, fearing that revealing it would cause panic.

## Chapter 8 outline
1. Sarah begins to notice strange occurrences around town, like unusual weather patterns and eerie shadows.
2. She consults the journal again and discovers a passage that seems to describe unusual event patterns as signs of "The Shadow's" return.
3. Sarah's friend Mark, who is a local historian, suggests that they investigate further by visiting the town's library.
4. At the library, they find more texts related to the secret society and "The Shadow."
5. Mark went home to rest after a long day, where he met his mother. 
<example_chapter_outline>

<example_claim>
Sarah decides to keep the discovery to herself, which reveals the challenges in Mark and Sarah's friendship.
</example_claim>

<example_verification>
<reasoning>
1. Analysis of atomic parts:
    a. Sarah decides to keep the discovery to herself: This part is grounded in event 4 of Chapter 3.
    b. "reveals the challenges in Mark and Sarah's friendship": This part is a subjective interpretation and not explicitly grounded in the summary or outline.
2. Analysis of the relationship between atomic parts:
    a. The relationship is not meaningful, as there is no direct connection between Sarah keeping the discovery to herself and revealing challenges in Mark and Sarah's friendship.
    b. The relationship is a subjective interpretation and not explicitly grounded in the summary or outline.
    c. The relationship does not contradict any information in the summary or outline.
Since 1b., 2a., and 2b. are not satisfied, the claim is INVALID.
</reasoning>
<result>
INVALID
</result>

<example_claim>
Sarah finds an old journal in the room, which prompted Mark to go home and meet his mother.
</example_claim>
<example_verification>
<reasoning>
1. Analysis of atomic parts:
    a. Sarah finds an old journal in the room: This part is grounded in event 2 of Chapter 3.
    b. Mark went home to rest after a long day, where he met his mother: This part is grounded in event 5 of Chapter 8.
\end{lstlisting} 
\end{tcolorbox}
\caption{Prompt for verifying claims with GPT-4o (Part 5)}
\label{fig:verification-prompt-5}
\end{figure*}

\begin{figure*}[htbp]
\centering
\begin{tcolorbox}[colback=gray!5!white, colframe=black, title=Prompt for Verifying Claims with GPT-4o (Part 6)]
\lstset{
    basicstyle=\ttfamily\footnotesize,
    breaklines=true,
    frame=none,
    xleftmargin=0pt,
    framexleftmargin=0pt,
    columns=fullflexible,
    tabsize=1,
    breakindent=0pt,
    breakautoindent=false,
    postbreak=\space,
    showstringspaces=false,
}
\begin{lstlisting}[firstline=161, lastline=200, language=Markdown]
2. Analysis of the relationship between atomic parts:
    a. The relationship is not meaningful, as there is no direct connection between Sarah finding the journal and Mark going home to meet his mother.
    b. The relationship is a subjective interpretation and not explicitly grounded in the summary or outline.
    c. The relationship does not contradict any information in the summary or outline.
Since 2a. and 2b. are not satisfied, the claim is INVALID.
</reasoning>
<result>
INVALID
</result>
</example_verification>
</example_2>

Now, it's your turn to verify the claim based on the provided book summary, chapter outline, and claim. Present your response in the following format:
<verification>
<reasoning>
[Provide your detailed reasoning here, explaining why the claim is or is not meaningful and coherent with the book summary and chapter outline.]
</reasoning>
<result>
[State whether the claim is VALID or INVALID. Use VALID if the claim portrays a meaningful relationship and is coherent with the book summary and chapter outline. Use INVALID if it does not.]
</result>
</verification>

Remember to base your verification solely on the information provided in the book summary, chapter outline, and the claim itself. Verify that the relationship makes sense and is objectively valid. Do not introduce external information or make assumptions beyond what is given.
\end{lstlisting} 
\end{tcolorbox}
\caption{Prompt for verifying claims with GPT-4o (Part 6)}
\label{fig:verification-prompt-6}
\end{figure*}

\begin{figure*}[htbp]
\centering
\begin{tcolorbox}[colback=gray!5!white, colframe=black, title=Prompt for Generating Book-level Claims in \naive]
\lstset{
    basicstyle=\ttfamily\footnotesize,
    breaklines=true,
    frame=none,
    xleftmargin=0pt,
    framexleftmargin=0pt,
    columns=fullflexible,
    tabsize=1,
    breakindent=0pt,
    breakautoindent=false,
    postbreak=\space,
    showstringspaces=false,
}
\begin{lstlisting}[language=Markdown]
Your task is to create factual statements that incorporate multiple events from multiple chapters of a book. These factual statements must be objective and specific, grounded in the involved chapters, and consistent with the entire book. The included events must be specific. The facts cannot contain any interpretations, speculations, or subjective statements. In addition to each fact, provide a minimally corrupted version of the fact, which sounds plausible but is wrong based on the entire book.
To create valid facts, follow these guidelines step-by-step:
1. Carefully read through the book.
2. In a <brainstorm> section: 
    - Identify different events that are strongly related to one another in multiple chapters.
    - Consider the relationship between these events, but DO NOT include this brainstorming in the resulting fact. If a meaningful relationship is found, move on to the next step; otherwise, proceed to the next fact.
3. Formulate your facts based on this analysis. Ensure that your facts:
    - Contain a single sentence
    - Are coherent with the entire book
    - Include multiple detailed and strongly related events from multiple chapters in the book
    - Are self-contained and independent of other facts
    - No part of the fact should be subjective interpretations or speculations
    - Do not contradict or duplicate existing facts
    - Do not contain chapter information (e.g. "In Chapter x, ...")
    - Do not contain quotes from the book. 
4. Formulate a minimally corrupted version of each fact. 
    - Only one aspect of the fact, such as an atomic detail or the relationship between details, should be altered.
    - The corrupted fact should sound plausible but be clearly wrong based on the entire book.
    - The corrupted fact should be similar with the original fact in terms of length and sentence structure to make it harder to verify solely based on surface-level features. 
First, read the book: 
<book>{book}</book>
Finally, review the existing facts:
<existing_facts>{existing_facts}</existing_facts>

Now, generate as many valid facts as possible based on the provided book. Return "No meaningful fact" if there is no valid fact. Present your facts in the following format:

<facts>
<fact_1>
<brainstorm>[Your brainstorm notes here]</brainstorm>
Fact: [Your fact here]
Fact Reasoning: [Your explanation here (including the chapter associated with each event)]
Source: [Chapters involved]
Corrupted Fact: [Your corrupted fact here]
Corrupted Fact Reasoning: [Your explanation here]
</fact_1>
[Continue with additional facts...]
</facts>
Remember to create facts that are objectively valid, coherent with the book, and demonstrate strong, meaningful relationships between the events from multiple chapters. No part of the fact can include subjective interpretations or generalizations. The relationship must be meaningful. The included events must be SPECIFIC and DETAILED!!!!!
\end{lstlisting}
\end{tcolorbox}
\caption{Prompt for generating book-level claims in \naive.}
\label{fig:book-alt}
\end{figure*}

\begin{figure*}[htbp]
\centering
\begin{tcolorbox}[colback=gray!5!white, colframe=black, title=Prompt for Generating Chapter-level Claims in \naive]
\lstset{
    basicstyle=\ttfamily\footnotesize,
    breaklines=true,
    frame=none,
    xleftmargin=0pt,
    framexleftmargin=0pt,
    columns=fullflexible,
    tabsize=1,
    breakindent=0pt,
    breakautoindent=false,
    postbreak=\space,
    showstringspaces=false,
}
\begin{lstlisting}[language=Markdown]
Your task is to create factual statements that incorporate multiple events from Chapter {chapter} of a book. These factual statements must be objective and specific, grounded in the specific chapter, and consistent with the entire book. The included events must be specific. The facts cannot contain any interpretations, speculations, or subjective statements. In addition to each fact, provide a minimally corrupted version of the fact, which sounds plausible but is wrong based on the entire book.
To create valid facts, follow these guidelines step-by-step:
1. Carefully read through the book.
2. In a <brainstorm> section: 
    - Identify different events that are strongly related to one another in a single chapter.
    - Consider the relationship between these events, but DO NOT include this brainstorming in the resulting fact. If a meaningful relationship is found, move on to the next step; otherwise, proceed to the next fact.
3. Formulate your facts based on this analysis. Ensure that your facts:
    - Contain a single sentence
    - Are coherent with the entire book
    - Include multiple detailed and strongly related events from a single chapter in the book
    - Are self-contained and independent of other facts
    - No part of the fact should be subjective interpretations or speculations
    - Do not contradict or duplicate existing facts
    - Do not contain chapter information (e.g. "In Chapter x, ...")
    - Do not contain quotes from the book. 
4. Formulate a minimally corrupted version of each fact. 
    - Only one aspect of the fact, such as an atomic detail or the relationship between details, should be altered.
    - The corrupted fact should sound plausible but be clearly wrong based on the entire book.
    - The corrupted fact should be similar with the original fact in terms of length and sentence structure to make it harder to verify solely based on surface-level features. 
First, read the book: 
<book>{book}</book>
Finally, review the existing facts:
<existing_facts>{existing_facts}</existing_facts>
Now, generate as many valid facts as possible based on the provided chapter. Return "No meaningful fact" if there is no valid fact. Present your facts in the following format:
<facts>
<fact_1>
<brainstorm>[Your brainstorm notes here]</brainstorm>
Fact: [Your fact here]
Fact Reasoning: [Your explanation here (including the chapter associated with each event)]
Source: [Chapter involved]
Corrupted Fact: [Your corrupted fact here]
Corrupted Fact Reasoning: [Your explanation here]
</fact_1>
[Continue with additional facts...]
</facts>
Remember to create facts that are objectively valid, coherent with the entire book, and demonstrate strong, meaningful relationships between the events from Chapter {chapter}. No part of the fact can include subjective interpretations or generalizations. The relationship must be meaningful. The included events must be SPECIFIC and DETAILED!!!!!
\end{lstlisting}
\end{tcolorbox}
\caption{Prompt for generating chapter-level claims in \naive.}
\label{fig:chapter-alt}
\end{figure*}

\subsection{Using DeepSeek-Distill to measure CoT groundedness} 
\label{appendix:deepseek-cot}
We evaluate the model on 66 annotated claims from \S\ref{data:claims_generation} and measure its agreement with human annotations (Table \ref{tab:judge_agreement}). Among the models tested, DeepSeek-Distill aligns most closely with human judgments, with only one instance of disagreement, outperforming other models like GPT-4o (10 disagreements) and LLaMA-3.1-70B-Instruct (3 disagreements). Although Llama-70B performs comparably, it fails to provide clear explanations for its decisions and instead generating generic reasoning messages that lack specificity to samples. Therefore, we use DeepSeek-Distill to measure CoT groundedness in our dataset.

\begin{table}[htbp]
    \small
    \centering
    \begin{tabular}{lc}
        \toprule
        Judge Models & \% Agreement \\
        \midrule
        GPT-4o & 84.8\% \\
        Llama-3.1-70b-Instruct & 95.5\% \\
        DeepSeek-Distill-Llama-70B & 98.5\%\\
         \bottomrule
    \end{tabular}
    \caption{Percentage of times LLM judges for chain of thought groundedness agree with our manual annotation over 66 samples in Section \ref{data:claims_generation}.}
    \label{tab:judge_agreement}
\end{table}

%% file: sections/A-training.tex
\section{Training}

\subsection{Codebases} To fine-tune models in the Llama family, we adopt the ProLong codebase,\footnote{\href{https://github.com/princeton-nlp/ProLong}{https://github.com/princeton-nlp/ProLong}} which integrates PyTorch~\citep{paszke2019pytorchimperativestylehighperformance} and Hugging Face~\citep{wolf2020huggingfacestransformersstateoftheartnatural} for model training, FlashAttention-2~\citep{dao2023flashattention2fasterattentionbetter} for efficient attention computation, and DeepSpeed Ulysses~\citep{jacobs_deepspeed_2023} for sequence parallelism, enabling training across 8 A100 GPUs. For fine-tuning \qweninst\ (Qwen-Instruct), we use the 360-LlamaFactory codebase,\footnote{\href{https://github.com/Qihoo360/360-LLaMA-Factory}{https://github.com/Qihoo360/360-LLaMA-Factory}} a modification of Llama-Factory\footnote{\href{https://github.com/hiyouga/LLaMA-Factory}{https://github.com/hiyouga/LLaMA-Factory}} that incorporates sequence parallelism via zigzag ring attention~\citep{liu2023blockwise,liu2023ring}. We choose \prolongbase\ (ProLong-base) over \prolonginst\ (ProLong-instruct) based on a small fine-tuning experiment, where we fine-tune both ProLong-instruct and ProLong-base on 2K training examples. This experiment shows that ProLong-base outperforms ProLong-instruct by 61.6\% and 59.7\%, respectively. 

\begin{figure}[htbp]
    \centering
    \includegraphics[width=\linewidth]{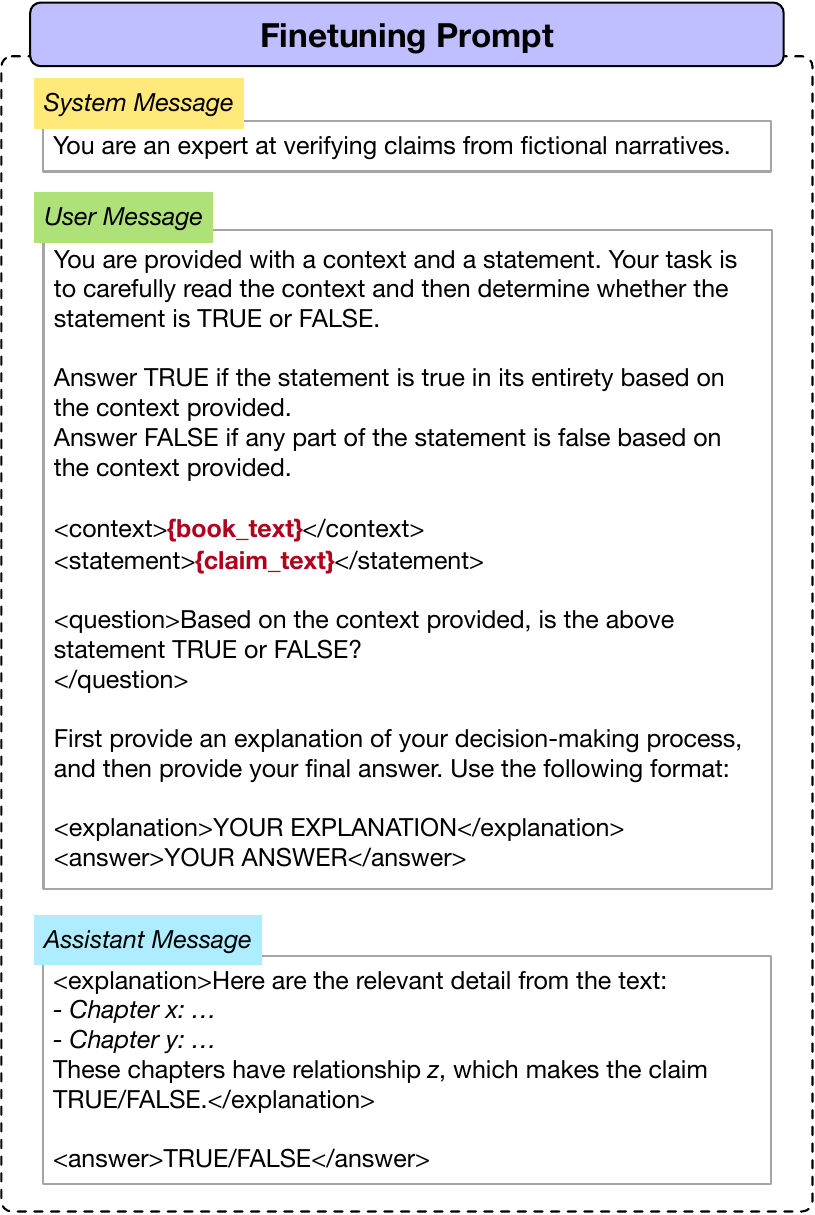}
    \caption{How we structure our fine-tuning prompts, which include system message, user message, and assistant message. Placeholders (colored in \textbf{\textcolor{lightcayenne}{light cayenne}}) will be replaced with actual text from the dataset. The contents between \texttt{<context>}, \texttt{<explanation>}, and \texttt{<answer>} tags are generated (Section \ref{data:claims_generation}).}
    \label{fig:sft_prompt}
\end{figure}
\subsection{Hyperparameter Tuning}
Table \ref{tab:hyperparam} summarizes the performance of each configuration from our hyperparameter tuning experiment on 100 samples from \pipeline's dev set.
\begin{table}[htbp]
\small
\centering
\begin{tabular}{@{}ccc@{}}
\toprule
Learning Rate & Batch Size &Dev Set Accuracy \\ \midrule
1e-5 & 16& 26\% \\
1e-6 & 16& \textbf{74}\% \\
1e-7 & 16& 71\% \\
1e-5 & 32& 34\% \\
1e-6 & 32& 73\% \\
1e-7 & 32& 69\% \\ \bottomrule
\end{tabular}
\caption{Hyperparameter tuning results. Each model is fine-tuned for 1 epoch and tested on a subset of 100 samples from our dev set.}
\label{tab:hyperparam}
\end{table}

%% file: sections/A-evaluation.tex
\section{Evaluation}

\begin{table*}[ht]
\centering
\footnotesize
\resizebox{\textwidth}{!}{%
\begin{tabular}{p{0.15\textwidth}p{0.25\textwidth}p{0.2\textwidth}p{0.2\textwidth}p{0.2\textwidth}}
\toprule
\multicolumn{1}{c}{\textsc{Category}} & \multicolumn{1}{c}{\textsc{Definition}} & \multicolumn{1}{c}{\textsc{Example Question}} & \multicolumn{1}{c}{\textsc{Example Ground-truth}} & \multicolumn{1}{c}{\textsc{Example Model Answer}} \\
\midrule
\addlinespace
\multicolumn{5}{c}{\textbf{ Cases that should be counted as incorrect }} \\
\addlinespace
\midrule
Wrong & The model's answer is evidently wrong. & What is the home Edna moves into in New Orleans called? & The pigeon house & The text does not provide information about Edna moving \\
\midrule
\addlinespace
\multicolumn{5}{c}{\textbf{ Cases that should be counted as correct }} \\
\addlinespace
\midrule
Full match & The model's answer perfectly matches the ground-truth answer. & Which among Annalisa, Seb, Peyton, and Gannonmarie is not Mrs. Bronwyn's child? & Peyton & Peyton \\
\midrule
Correct (Phrasing) & The model's answer is correct, but it has a different phrasing than the ground-truth answer. & How old is Felicity at the start of his narration? & Thirty & Felicity is 30 years old at \\
\midrule
Correct (Long) & The model's answer is correct, but it is longer than the ground-truth. & What kind of pet does Madame Bowen keep? & a cat & Madame Bowen keeps a cat as a pet. \\
\midrule
Correct (Short) & The model's answer is correct, but it is shorter than the ground-truth. & Why is Tasha Teigan out of jail? & He has been released on parole. & Paroled. \\
\midrule
\addlinespace
\multicolumn{5}{c}{\textbf{ Cases where correctness is ambiguous }} \\
\addlinespace
\midrule
Truncation & The model's answer has been heavily truncated, making it impossible to tell the correctness of the answer. & What is to be built in place of the Lars home on Wickham Place? & Flats & The house on Wickham Place is to be replaced \\
\bottomrule
\end{tabular}}
\caption{Taxonomy from our analysis on the $\infty$Bench QA outputs of Qwen2.5-7B-Instruct and \qwenftbalanced. Example model outputs are from \qwenftbalanced\ except the one for ``Correct (Short)", which is from Qwen2.5-7B-Instruct (all generated under the default setup where maximum output tokens is set to 10).}
\label{tab:infbench-taxonomy}
\end{table*}

\begin{table}[ht]
\centering
\footnotesize
\scalebox{1}{
\begin{tabular}{p{0.15\textwidth}p{0.10\textwidth}p{0.10\textwidth}}
\toprule
\multicolumn{1}{c}{\textsc{Category}} & \multicolumn{1}{c}{\textsc{Qwen-Inst}} & \multicolumn{1}{c}{\textsc{Qwen-\texttt{BC}}} \\
\midrule
\addlinespace
\multicolumn{3}{c}{\textbf{ Cases that should be counted as incorrect }} \\
\addlinespace
\midrule
Wrong & 41 & 26 \\
\midrule
\addlinespace
\multicolumn{3}{c}{\textbf{ Cases that should be counted as correct }} \\
\addlinespace
\midrule
Full match & 17 & 4 \\
\midrule
Correct (Phrasing) & 3 & 7 \\
\midrule
Correct (Long) & 13 & 18 \\
\midrule
Correct (Short) & 2 & 0 \\
\midrule
\addlinespace
\multicolumn{3}{c}{\textbf{ Cases where correctness is ambiguous }} \\
\addlinespace
\midrule
Truncation & 24 & 45 \\
\bottomrule
\end{tabular}}
\caption{Raw counts of taxonomy categories for \qweninst\ and \qwenftbalanced, with outputs generated using the default maximum length of 10 tokens.}
\label{tab:infbench-counts}
\end{table}

\begin{table*}[ht]
\centering
\footnotesize
\resizebox{\textwidth}{!}{%
\begin{tabular}{p{0.2\textwidth}p{0.15\textwidth}p{0.2\textwidth}p{0.1\textwidth}p{0.3\textwidth}}
\toprule
\multicolumn{1}{c}{\textsc{Question}} & \multicolumn{1}{c}{\textsc{Ground-truth}} & \multicolumn{1}{c}{\textsc{Model Answer}} & \multicolumn{1}{c}{\textsc{ROUGE F1}} & \multicolumn{1}{c}{\textsc{Explanation}} \\
\midrule
How old is Felicity at the start of his narration? & Thirty & Felicity is 30 years old at & 0 & The model is correct, but it uses the numerical form of the number. \\
\midrule
What gender does Harris predict Cal will be? & MALE & Harris predicts that Cal will be a boy. & 0 & The model is correct, but it phrases it differently, resulting in no word overlap. \\
\midrule
When is Jarod's birthday? & NOVEMBER 9 & Jarod's birthday is on November 16 & 0.22 & The model is completely wrong, but it gets the same score as the model answer in the row below, which contains a correct answer. \\
\midrule
In which state is Gopher Prairie located? & Minnesota & Gopher Prairie is located in Minnesota. This is & 0.22 & The model is correct, but it gets the same score as the wrong model answer above, just because the output is much longer than the ground-truth. \\
\bottomrule
\end{tabular}}
\caption{Examples showing that ROUGE-F1 is an unreliable metric.}
\label{tab:infbench-rouge-examples}
\end{table*}

\subsection{Configuration for $\infty$Bench QA Evaluation} \label{appendix:infbench-metric}

In HELMET \citep{yen2024helmet}, for the \texorpdfstring{$\infty$}Bench QA task, the default configuration sets the output maximum length to 10 tokens and uses ROUGE F1 \citep{lin-2004-rouge} as the evaluation metric. Upon closer examination of the outputs from both models, we identify critical flaws in the default setup. These findings eventually motivate us to remove the maximum length restriction and adopt the LLM-as-a-judge evaluation approach using GPT-4o as the judge. Below, we provide more details on our analysis.

\paragraph{Setting max output length to 10 tokens frequently cause truncations:} In Table \ref{tab:infbench-taxonomy}, we show the taxonomy we derive from our analysis. Here, we define a truncation to be when a model's response is heavily cut off, \textit{making it impossible to determine the correctness of the output}. Out of all 100 evaluated examples, Qwen2.5-7B-Instruct's outputs get truncated 24 times, while \qwenftbalanced's outputs get truncated 45 times. After removing the 10-token maximum length restriction\footnote{Without the maximum length limit, Qwen2.5-7B-Instruct's outputs are on average 20.9 tokens long, and \qwenftbalanced's outputs have 25.8 tokens on average.}, we observe that 9 of the 24 previously truncated outputs from \qweninst should be counted as correct. For \qwenftbalanced, this correction is even more significant, with 25 of the 45 truncated outputs being technically correct. We combine these numbers with numbers from the four rows in Table \ref{tab:infbench-counts} that indicate correctness, and find that Qwen2.5-7B-Instruct has an overall accuracy of 44\%, while \qwenftbalanced\ has 54\%.

\paragraph{ROUGE F1 is not a reliable metric:} If we use ROUGE F1 as the metric, Qwen2.5-7B-Instruct achieves a score of 27.4, while \qwenftbalanced achieves a score of 18.0. This result sharply contrasts with the accuracies we obtain in the preceding paragraph, and does not reflect the actual performance of the models. Lots of prior work have shown that ROUGE correlates poorly with human judgment \citep{goyal2023newssummarizationevaluationera, chang_booookscore_2024}. Our manual analysis reveals that this metric is overly sensitive to length, and does not capture the correctness of the model outputs. We show several examples in Table \ref{tab:infbench-rouge-examples}.

%% file: sections/A-results.tex
\section{Results}
\subsection{Impact of chapter distance and book length on test set performance}
\begin{figure}[htbp]
    \centering
    \includegraphics[width=\linewidth]{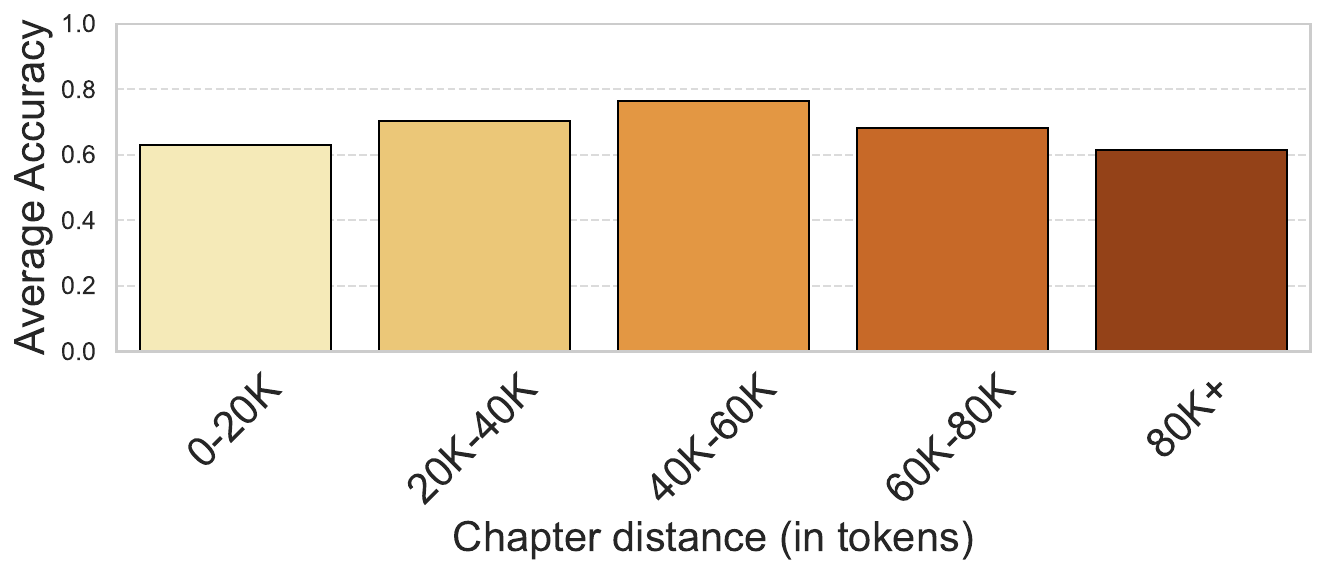}
    \caption{Accuracy of \pipeline-Prolong-balanced on the test set (book-level claims), grouped by the distance (in tokens) between source events in each claim.}
    \label{fig:accuracy_chapter}
\end{figure}

Figure \ref{fig:accuracy_chapter} shows that test set accuracy peaks when the distance between chapters in a claim is around 40--60K tokens (roughly the midpoint of a book). When that gap shrinks below or stretches beyond 60K tokens, performance dips by about 10\%, leaving no definitive pattern beyond this sweet spot.

\begin{figure}[htbp]
    \centering
    \includegraphics[width=\linewidth]{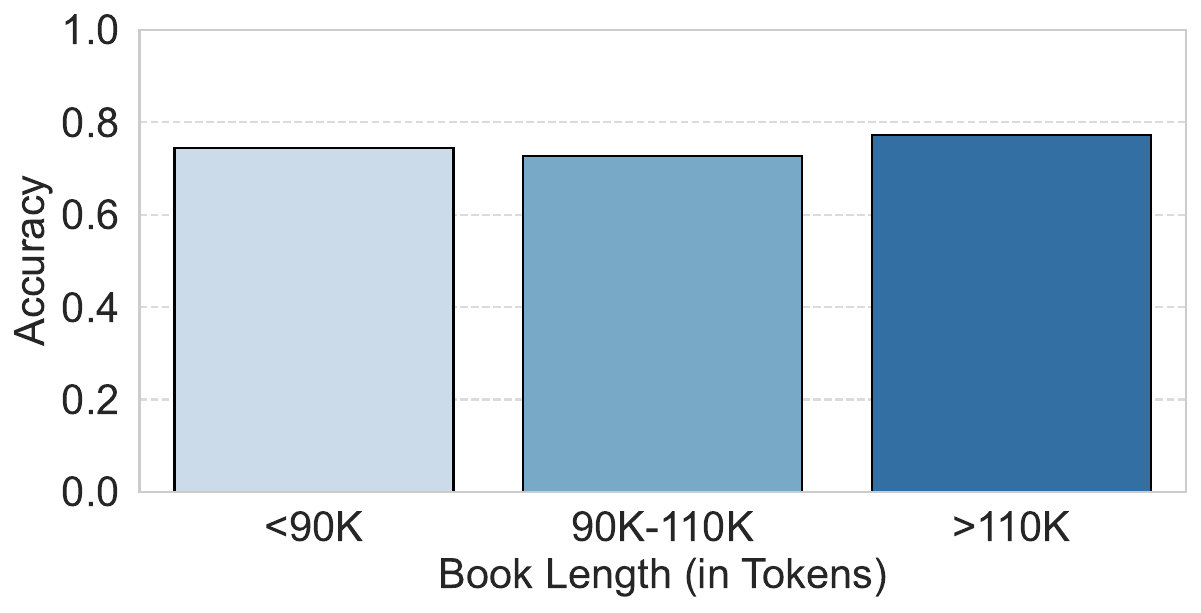}
    \caption{\prolongftbalanced's performance on test set, grouped by the number of tokens in each book.}
    \label{fig:accuracy_book}
\end{figure}

We also find that overall book length does not strongly influence accuracy, except in cases where the text exceeds 110K tokens. In these longer works, accuracy is about 5\% higher than in shorter books, as shown in Figure~\ref{fig:accuracy_book}. While this slight edge may hint at advantages in more expansive narratives, the model’s broader performance remains steady across most book lengths.

\begin{figure}[htbp]
    \centering
    \includegraphics[width=\linewidth]{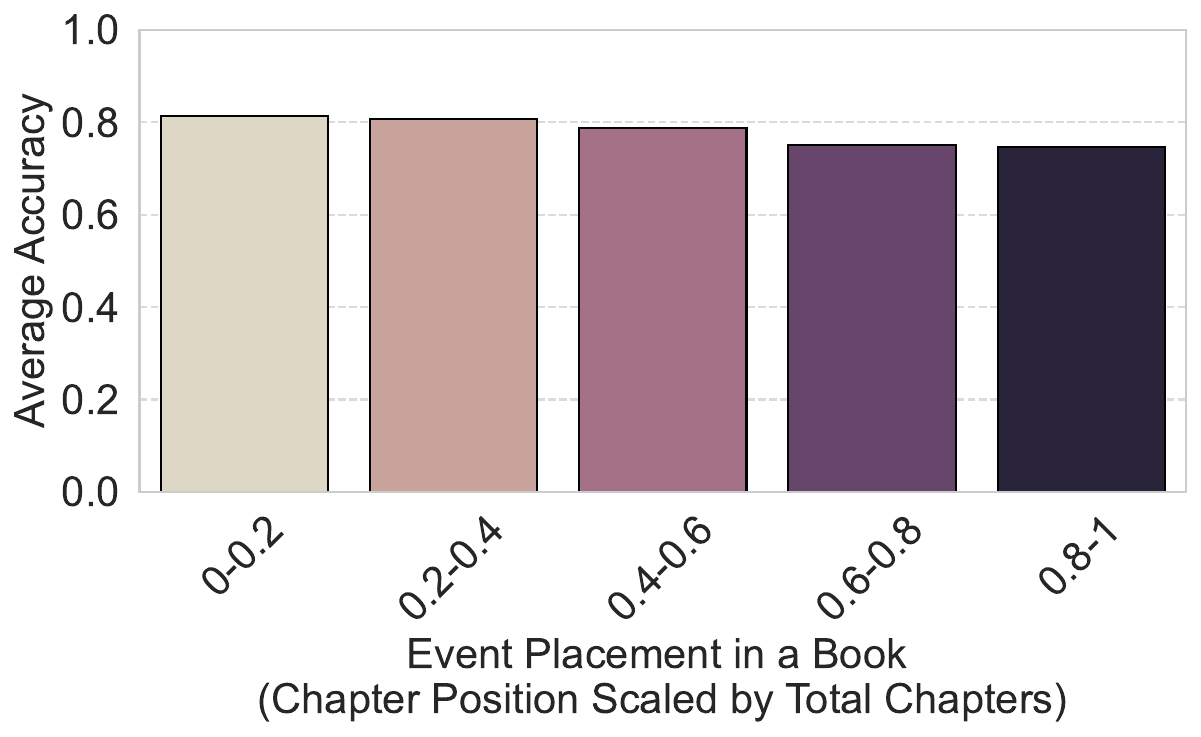}
    \caption{Accuracy of \prolongftbalanced\ on the test set (chapter-level claims), grouped by the event placement in the book (0-0.2 includes events are at the beginning, while 0.8-1 includes events towards the end). }
    \label{fig:accuracy_event}
\end{figure}

We finally examine the possible effect of event placement on \prolongftbalanced's performance on the test set. Interestingly, there is no strong "lost-in-the-middle" effect regarding event placement in the book \citep{liu2024lost}. As shown in Figure \ref{fig:accuracy_event}, accuracy is usually the highest when the claim involves events that appear at the beginning (0-0.4, around 82\%) rather than at the end of the book (0.8-1, around 78\%). 

\subsection{False claim error analysis} \label{appendix:error-analysis}

In Table \ref{tab:definitions}, we provide detailed definitions for each category from the false claim error analysis in \S\ref{sec:error-analysis}. To explore instances where fine-tuned models still struggle, we conduct an in-depth analysis of \qwenftbalanced\ outputs. Of the 1,000 book-level claims in the test set, the model fails to verify 37 true claims and 97 false claims. This pattern is consistent with findings from NoCha \citep{karpinska_one_2024}, which highlight that models tend to have greater difficulty verifying false claims. Notably, in 95 cases, the model successfully validates the true claim but fails to validate the corresponding false claim. This raises an important question: \textit{What specific perturbations make a false claim appear true to the model?} 

Through careful manual analysis, we derive a taxonomy of such perturbations and present them in Table \ref{tab:error-analysis-dist}. The most frequent perturbations are changes to events (43.2\%) and people (31.6\%), such as altering actions or misattributing roles. Less frequent but notable are modifications to objects (15.8\%), locations (13.7\%), time (6.3\%), and affect (4.2\%). All these perturbations introduce plausible-sounding variations that the model may struggle to detect without fully understanding the narrative.

\begin{table*}[ht]
\centering
\scalebox{0.87}{
\begin{tabular}{p{0.1\textwidth}p{0.9\textwidth}}
\toprule
\multicolumn{1}{c}{\textsc{Category}} & \multicolumn{1}{c}{\textsc{Definition}} \\
\midrule
Event & Refers to the alteration or misrepresentation of the actions, occurrences, or processes described in a claim. \\
\midrule
Person & Involves substituting or misattributing individuals involved in a claim. \\
\midrule
Object & Concerns the manipulation or substitution of physical items or artifacts mentioned in a claim. \\
\midrule
Location & Relates to changing or misrepresenting the places where events occur. \\
\midrule
Time & Pertains to the sequencing or timing of events being distorted or swapped. \\
\midrule
Affect & Deals with altering the emotional state, attitude, or disposition described in a claim. \\
\bottomrule
\end{tabular}}
\caption{Definitions for each category of perturbations that cause a false claim to be misclassified as true in the error analysis in \S\ref{sec:error-analysis}.}
\label{tab:definitions}
\end{table*}

\subsection{Full results on LM Harness and HELMET}
\label{appendix:full_results}
Table \ref{tab:lm-harness} shows the results of all models on popular short-form benchmarks. Overall, our fine-tuned models, especially \qwenftbalanced, do not degrade that significantly from the baseline models even though it has been fine-tuned on longer data. 
Table \ref{tab:helmet} shows the results of HELMET on recall, RAG, passage re-ranking, and retrieval tasks. Overall, fine-tuned models achieve synthetic recall and RAG scores comparable to the baseline models, while generally delivering improved re-ranking and more robust ICL performance.
\begin{table*}[htpb]
    \centering
    \resizebox{\textwidth}{!}{%
    \begin{tabular}{lccccccccc}
        \toprule
        Models & IFEval & BBH & Math lvl5 & GPQA & MMLU-Pro & Arc-Challenge & GSM8K & HellaSwag & WinoGrande \\
        \midrule
        \llamainst\ & \textbf{59.35} & 50.93 & 12.81 & 31.96 & 37.77 & 51.54 & 75.06 & 59.05 & 74.19 \\
        \qweninst\ & 54.00 & 54.60 & \textbf{24.80} & 33.40 & 43.80 & 53.20 & 77.70 & 61.80 & 69.20 \\
        Prolong-instruct-noft & 58.87 & 49.86 & 5.28 & 29.35 & 32.43 & \textbf{58.36} & 68.06 & \textbf{80.75} & \textbf{74.43} \\
        \midrule
        \qwenftbalanced\ & 50.65 & \textbf{55.50} & 22.51 & \textbf{33.82} & \textbf{44.49} & 53.84 & \textbf{78.32} & 61.71 & 69.06 \\
        \prolongftbalanced\ & 7.91 & 48.03 & 5.42 & 27.88 & 32.35 & 50.68 & 60.80 & 60.44 & 73.24 \\
        \llamaftbalanced\ & 45.43 & 50.02 & 12.77 & 30.59 & 37.55 & 53.50 & 74.91 & 78.65 & 73.40 \\
        \prolongwp\ & 11.75 & 47.32 & 3.59 & 30.73 & 26.29 & 50.51 & 39.04 & 76.36 & 70.64 \\
        \prolongftbook\ & 6.39 & 49.31 & 5.53 & 29.47 & 32.38 & 54.78 & 62.02 & 79.27 & 72.93 \\
        \prolongftchapter\ & 4.59 & 49.64 & 5.63 & 29.77 & 32.35 & 54.27 & 61.22 & 79.14 & 74.27 \\
        \bottomrule
    \end{tabular}
    }
    \caption{Performance on popular short-form benchmarks (evaluated using Language Model Evaluation Harness).}
    \label{tab:lm-harness}
    \vspace{-0.1in}
\end{table*}

\begin{table*}[ht]
  \centering
  \resizebox{\textwidth}{!}{%
  \small
  \begin{tabular}{l*{13}{c}}
    \toprule
    & \multicolumn{4}{c}{Synthetic Recall (Ruler)} 
    & \multicolumn{3}{c}{RAG} 
    & \multicolumn{1}{c}{Re-ranking} 
    & \multicolumn{5}{c}{ICL} \\
    \cmidrule(lr){2-5} \cmidrule(lr){6-8} \cmidrule(lr){9-9} \cmidrule(lr){10-14}
    Model & 
      niah\_mk\_2 & recall & 
      niah\_mk\_3 & recall & 
      niah\_mv & recall & 
      \makebox[1.5cm][c]{json\_kv} & 
      \makebox[1.5cm][c]{nqh} & 
      \makebox[1.5cm][c]{triviaqa} & 
      \makebox[1.5cm][c]{hotpotqa} & 
      \makebox[1.8cm][c]{msmarco} & 
      \makebox[1.5cm][c]{trec\_coarse} & 
      \makebox[1.5cm][c]{trec\_fine} \\
    \midrule
    \llamainst       & 98  & 88     & 78.75 & 96   & 48.17 & 80.67 & 56   & 13.66 & 73  & 72  & 91  & 91  & 88 \\
    \qweninst        & \textbf{100.0} & 98.0  & \textbf{83.3}  & 98.8 & 20.3  & 47.3  & 24.0 & 0     & 78.0 & 20.0 & 6.0 & 11.0 & 7.0 \\
    \prolonginst     & 98.0  & 98.0  & 46.8  & 99.3 & \textbf{54.3}  & 91.3  & \textbf{57.7} & 25.0  & 86.0 & 59.0 & \textbf{92.0} & 94.0 & 89.0 \\
    \midrule
    \prolongftbalanced & 99    & 92    & 27.75 & 99   & 52.5  & \textbf{92}    & 51.33& \textbf{27.50} & \textbf{92}   & 72   & 90   & 94   & \textbf{90} \\
    \qwenftbalanced   & 81    & 45    & 45    & 45   & 38.16 & 66.5  & 36   & 3.17  & 73   & 52   & 87   & 78   &  \\[0.5ex]
    \llamaftbalanced  & 98    & \textbf{99}    & 26    & \textbf{100}  & 48.5  & 85.17 & 55.67& 22.90 & 87   & \textbf{81}   & 90   & \textbf{95}   & 86 \\
    \prolongwp        & 99    & 87    & 31.75 & \textbf{100}  & 50.5  & 89.17 & 51.33& 18.30 & 91   & 65   & 91   & \textbf{95}   & 88 \\
    \bottomrule
  \end{tabular}%
  }
  \caption{Performance on HELMET for recall, RAG, passage re-ranking, and retrieval tasks. Fine-tuned models achieve synthetic recall and RAG scores comparable to the baseline models, while generally delivering improved re-ranking and more robust ICL performance.}
  \label{tab:helmet}
\end{table*}

\subsection{Performance of \prolongbase\ on claim verification and narrative understanding benchmarks}
Table \ref{tab:prolong-base-acc} shows accuracy of \prolongbase on long-context reasoning and narrative understanding benchmarks. Even though \prolongbase's test set performance is much worse than \prolonginst, performance on other narrative understanding tasks is comparable between the two models.
\begin{table}[htpb]
    \centering
    \small
    \begin{tabular}{cccc}
        \toprule
         \pipeline-Test & NarrativeQA & MuSR & $\infty$Bench QA\\
        \midrule
        23.9\% & 46.0\% & 39.8\% & 42.5\%\\
        \bottomrule
    \end{tabular}
    \caption{Performance of \prolongbase\ on long-context reasoning and narrative understanding benchmarks. Even though \prolongbase's test set performance is much worse than \prolonginst, performance on other narrative understanding tasks is comparable between the two models.}
    \label{tab:prolong-base-acc}
    \vspace{-0.1in}
\end{table}

\begin{table}[htbp]
    \centering
    \small
    \begin{tabular}{lcc}
        \toprule
        \multicolumn{1}{l}{} & \multicolumn{1}{c}{ProLong-\texttt{CP}-book} & \multicolumn{1}{c}{ProLong-\texttt{CP}-chap} \\
        \midrule
        Test-book     & 74.8\%  & 78.2\% \\
        Test-chapter  & 75.2\%  & 80.2\% \\
        \midrule
        Overall       & 75.0\%  & 79.2\% \\ 
        \bottomrule
    \end{tabular}
    \caption{Test set performance of models trained exclusively on either book-level claims or chapter-level claims, with accuracy measured for book-level, chapter-level, and overall claims. \texttt{CP} stands for \pipeline.}
    \label{tab:chapter_vs_book}
    \vspace{-0.1in}
\end{table}

\begin{table}[htbp]
\centering
\footnotesize
\begin{tabular}{lc}
\toprule
Models   & Groundedness \\ \midrule
\qweninst\     & 11.9\%\\
\llamainst\       & 16.8\%\\
\prolonginst\ & 19.6\%\\ \midrule
\qwenftbalanced\          & 67.1\%    \\
\llamaftbalanced\         & 75.9\%\\
\prolongftbalanced\       & \textbf{80.6}\%\\
\bottomrule
\end{tabular}
\caption{Percentage of grounded chain of thoughts being generated by baseline and fine-tuned models. Our fine-tuned models generate much more grounded chain of thoughts.}
\label{tab:cot-groundedness}
\vspace{-0.1in}
\end{table}

\subsection{Performance of retrieval-augmented baselines on \pipeline-test}
To better isolate the contribution of \pipeline, we present the performance of a retrieval-augmented approach on \pipeline-test. Specifically, we report results of an approach where we use BM25 \citep{robertson1995okapi} to retrieve the top 50 relevant book passages (each no longer than 256 words) for a given claim and prompting our original baselines with these passages instead of the full book text.

\begin{table}[h]
\centering
\begin{tabular}{l c}
\toprule
\textbf{Model} & \textbf{CLIPPER-test (\%)} \\
\midrule
Llama-3.1-8B-Instruct & 27.9 \\
ProLong-512K-8B-Instruct & 34.5 \\
Qwen2.5-7B-Instruct & 51.0 \\
\midrule
BM25 + Llama-3.1-8B-Instruct & 36.45 \\
BM25 + ProLong-512K-8B-Instruct & 40.0 \\
BM25 + Qwen2.5-7B-Instruct & 36.0 \\
\midrule
\llamaftbalanced & 76.0 \\
\prolongftbalanced & 75.0 \\
\qwenftbalanced & 73.9 \\
\bottomrule
\end{tabular}
\caption{Performance on CLIPPER-test for various models.}
\label{tab:rag_basline}
\end{table}

As seen in \autoref{tab:rag_basline}, these RAG baselines (denoted by BM25 + model name) outperform our original LLaMA and ProLong baselines (+9\%), but not the Qwen baseline (-15\%). Compared to our \pipeline\ models, however, these RAG baselines still lag by 22-50\%. It is important to note that RAG approaches do not consistently outperform long-context models in long-form claim verification. For instance, \citet{karpinska_one_2024} and \citet{kim_fables_2024} benchmark a similar RAG setup where GPT-4o is provided only with BM25-retrieved passages. As shown in Table 3 of \citet{karpinska_one_2024}, the RAG versions (k = 5, 25, 50) consistently underperform the setting where GPT-4o has no retrieval support.

\subsection{Performance margin of \pipeline}
\autoref{tab:stats} lists statistical test results for the performance reported in Table~\ref{tab:main-result}. For \pipeline-test, NoCha, and MuSR, which return binary True/False predictions, we use McNemar’s test~\citet{mcnemar1947note}. For NarrativeQA and InfiniBenchQA, which return ordinal scores ranging from 0 to 3, we use the Wilcoxon signed-rank test~\citet{wilcoxon1992individual}. 

Fine-tuning on \pipeline\ yields statistically significant improvements across all models on \pipeline-test and NoCha. For MuSR, both Qwen and LLaMA show significant gains, while ProLong does not. For $\infty$BenchQA, Qwen demonstrates a statistically significant improvement. For NarrativeQA, no models exhibit a significant improvement. 

While improvements on NarrativeQA, MuSR, and $\infty$BenchQA are modest, these results represent performance on OOD tasks in our paper. NarrativeQA and $\infty$BenchQA focus on question answering over narrative contexts, while MuSR consists of algorithmically generated reasoning problems. Therefore, significant performance gains on these tasks would be nice to have, not expected.

\begin{table*}[t]
\centering
\footnotesize
\setlength{\tabcolsep}{4pt}
\begin{adjustbox}{max width=\textwidth}
\begin{tabular}{lccccc}
\toprule
\textbf{Baseline Models} & \textbf{CLIPPER-test ($\chi^2$)} & \textbf{NoCha ($\chi^2$)} & \textbf{NarrativeQA (Wilcoxon)} & \textbf{MuSR ($\chi^2$)} & \textbf{InfiniBenchQA (Wilcoxon)} \\
\midrule
Qwen2.5-7B-Instruct        & 174.0 & 65.0 & 105.0 & 38.0 & 2825.5 \\
Llama-3.1-8B-Instruct      &   0.0 & 52.0 & 205.5 & 36.0 & 2029.5 \\
ProLong-512K-8B-Instruct   &  82.0 & 54.0 & 156.0 & 90.0 & 4556.5 \\
\bottomrule
\end{tabular}
\end{adjustbox}
\caption{\label{tab:stats} Test statistics comparing fine-tuned and baseline models across benchmarks. For CLIPPER-test, NoCha, and MuSR we report McNemar's $\chi^2$ statistic; for NarrativeQA and InfiniBenchQA we report the Wilcoxon signed-rank statistic. \autoref{tab:pvalues} shows p-value corresponding to these test statistics.}
\end{table*}

\begin{table*}[t]
\centering
\footnotesize
\setlength{\tabcolsep}{4pt}
\begin{adjustbox}{max width=\textwidth}
\begin{tabular}{lccccc}
\toprule
\textbf{Baseline Models} & \textbf{CLIPPER-test} & \textbf{NoCha} & \textbf{NarrativeQA} & \textbf{MuSR} & \textbf{InfiniBenchQA} \\
\midrule
Qwen2.5-7B-Instruct        & 8.719614e$-$62 & 7.929521e$-$10 & 0.349212 & 0.000351 & 0.002420 \\
Llama-3.1-8B-Instruct      & 6.406666e$-$145 & 1.467807e$-$12 & 0.791416 & 0.003866 & 0.544319 \\
ProLong-512K-8B-Instruct   & 1.212688e$-$165 & 6.404008e$-$05 & 0.351747 & 0.316073 & 0.108244 \\
\bottomrule
\end{tabular}
\end{adjustbox}
\caption{\label{tab:pvalues} p-values for statistical significance (threshold $p<0.05$). See \autoref{tab:stats} for test statistics.}
\end{table*}

%% file: main.bbl
\begin{thebibliography}{71}
\providecommand{\natexlab}[1]{#1}
\providecommand{\url}[1]{\texttt{#1}}
\expandafter\ifx\csname urlstyle\endcsname\relax
  \providecommand{\doi}[1]{doi: #1}\else
  \providecommand{\doi}{doi: \begingroup \urlstyle{rm}\Url}\fi

\bibitem[AlKhamissi et~al.(2023)AlKhamissi, Verma, Yu, Jin, Celikyilmaz, and Diab]{alkhamissi_opt-r_2023}
Badr AlKhamissi, Siddharth Verma, Ping Yu, Zhijing Jin, Asli Celikyilmaz, and Mona Diab.
\newblock {OPT}-{R}: {Exploring} the {Role} of {Explanations} in {Finetuning} and {Prompting} for {Reasoning} {Skills} of {Large} {Language} {Models}.
\newblock In \emph{Proceedings of the 1st {Workshop} on {Natural} {Language} {Reasoning} and {Structured} {Explanations} ({NLRSE})}, pp.\  128--138, 2023.
\newblock \doi{10.18653/v1/2023.nlrse-1.10}.
\newblock URL \url{http://arxiv.org/abs/2305.12001}.
\newblock arXiv:2305.12001 [cs].

\bibitem[An et~al.(2024)An, Ma, Lin, Zheng, and Lou]{an2024makellmfullyutilize}
Shengnan An, Zexiong Ma, Zeqi Lin, Nanning Zheng, and Jian-Guang Lou.
\newblock Make your llm fully utilize the context, 2024.
\newblock URL \url{https://arxiv.org/abs/2404.16811}.

\bibitem[Bai et~al.(2024)Bai, Lv, Zhang, He, Qi, Hou, Tang, Dong, and Li]{bai_longalign_2024}
Yushi Bai, Xin Lv, Jiajie Zhang, Yuze He, Ji~Qi, Lei Hou, Jie Tang, Yuxiao Dong, and Juanzi Li.
\newblock {LongAlign}: {A} {Recipe} for {Long} {Context} {Alignment} of {Large} {Language} {Models}!, January 2024.
\newblock URL \url{http://arxiv.org/abs/2401.18058}.
\newblock arXiv:2401.18058 [cs].

\bibitem[Chang et~al.(2024)Chang, Lo, Goyal, and Iyyer]{chang_booookscore_2024}
Yapei Chang, Kyle Lo, Tanya Goyal, and Mohit Iyyer.
\newblock {BooookScore}: {A} systematic exploration of book-length summarization in the era of {LLMs}, April 2024.
\newblock URL \url{http://arxiv.org/abs/2310.00785}.
\newblock arXiv:2310.00785 [cs].

\bibitem[Chung et~al.(2022)Chung, Hou, Longpre, Zoph, Tay, Fedus, Li, Wang, Dehghani, Brahma, Webson, Gu, Dai, Suzgun, Chen, Chowdhery, Castro-Ros, Pellat, Robinson, Valter, Narang, Mishra, Yu, Zhao, Huang, Dai, Yu, Petrov, Chi, Dean, Devlin, Roberts, Zhou, Le, and Wei]{chung_scaling_2022}
Hyung~Won Chung, Le~Hou, Shayne Longpre, Barret Zoph, Yi~Tay, William Fedus, Yunxuan Li, Xuezhi Wang, Mostafa Dehghani, Siddhartha Brahma, Albert Webson, Shixiang~Shane Gu, Zhuyun Dai, Mirac Suzgun, Xinyun Chen, Aakanksha Chowdhery, Alex Castro-Ros, Marie Pellat, Kevin Robinson, Dasha Valter, Sharan Narang, Gaurav Mishra, Adams Yu, Vincent Zhao, Yanping Huang, Andrew Dai, Hongkun Yu, Slav Petrov, Ed~H. Chi, Jeff Dean, Jacob Devlin, Adam Roberts, Denny Zhou, Quoc~V. Le, and Jason Wei.
\newblock Scaling {Instruction}-{Finetuned} {Language} {Models}, December 2022.
\newblock URL \url{http://arxiv.org/abs/2210.11416}.
\newblock arXiv:2210.11416 [cs].

\bibitem[Dao(2023)]{dao2023flashattention2fasterattentionbetter}
Tri Dao.
\newblock Flashattention-2: Faster attention with better parallelism and work partitioning, 2023.
\newblock URL \url{https://arxiv.org/abs/2307.08691}.

\bibitem[Dao et~al.(2022)Dao, Fu, Ermon, Rudra, and Ré]{dao2022flashattentionfastmemoryefficientexact}
Tri Dao, Daniel~Y. Fu, Stefano Ermon, Atri Rudra, and Christopher Ré.
\newblock Flashattention: Fast and memory-efficient exact attention with io-awareness, 2022.
\newblock URL \url{https://arxiv.org/abs/2205.14135}.

\bibitem[DeepSeek-AI et~al.(2025)DeepSeek-AI, Guo, Yang, Zhang, Song, Zhang, Xu, Zhu, Ma, Wang, et~al.]{deepseekai2025deepseekr1incentivizingreasoningcapability}
DeepSeek-AI, Daya Guo, Dejian Yang, Haowei Zhang, Junxiao Song, Ruoyu Zhang, Runxin Xu, Qihao Zhu, Shirong Ma, Peiyi Wang, and 190 others.
\newblock Deepseek-r1: Incentivizing reasoning capability in llms via reinforcement learning, 2025.
\newblock URL \url{https://arxiv.org/abs/2501.12948}.

\bibitem[Ding et~al.(2023)Ding, Chen, Xu, Qin, Zheng, Hu, Liu, Sun, and Zhou]{ding_enhancing_2023}
Ning Ding, Yulin Chen, Bokai Xu, Yujia Qin, Zhi Zheng, Shengding Hu, Zhiyuan Liu, Maosong Sun, and Bowen Zhou.
\newblock Enhancing {Chat} {Language} {Models} by {Scaling} {High}-quality {Instructional} {Conversations}, May 2023.
\newblock URL \url{http://arxiv.org/abs/2305.14233}.
\newblock arXiv:2305.14233 [cs].

\bibitem[Dubey et~al.(2024)Dubey, Jauhri, Pandey, Kadian, Al-Dahle, Letman, Mathur, Schelten, Yang, Fan, et~al.]{dubey2024llama}
Abhimanyu Dubey, Abhinav Jauhri, Abhinav Pandey, Abhishek Kadian, Ahmad Al-Dahle, Aiesha Letman, Akhil Mathur, Alan Schelten, Amy Yang, Angela Fan, et~al.
\newblock The llama 3 herd of models.
\newblock \emph{arXiv preprint arXiv:2407.21783}, 2024.

\bibitem[Fan et~al.(2018)Fan, Lewis, and Dauphin]{fan-etal-2018-hierarchical}
Angela Fan, Mike Lewis, and Yann Dauphin.
\newblock Hierarchical neural story generation.
\newblock In Iryna Gurevych and Yusuke Miyao (eds.), \emph{Proceedings of the 56th Annual Meeting of the Association for Computational Linguistics (Volume 1: Long Papers)}, pp.\  889--898, Melbourne, Australia, July 2018. Association for Computational Linguistics.
\newblock \doi{10.18653/v1/P18-1082}.
\newblock URL \url{https://aclanthology.org/P18-1082/}.

\bibitem[Gao et~al.(2024{\natexlab{a}})Gao, Tow, Abbasi, Biderman, Black, DiPofi, Foster, Golding, Hsu, Le~Noac'h, Li, McDonell, Muennighoff, Ociepa, Phang, Reynolds, Schoelkopf, Skowron, Sutawika, Tang, Thite, Wang, Wang, and Zou]{eval-harness}
Leo Gao, Jonathan Tow, Baber Abbasi, Stella Biderman, Sid Black, Anthony DiPofi, Charles Foster, Laurence Golding, Jeffrey Hsu, Alain Le~Noac'h, Haonan Li, Kyle McDonell, Niklas Muennighoff, Chris Ociepa, Jason Phang, Laria Reynolds, Hailey Schoelkopf, Aviya Skowron, Lintang Sutawika, Eric Tang, Anish Thite, Ben Wang, Kevin Wang, and Andy Zou.
\newblock A framework for few-shot language model evaluation, 07 2024{\natexlab{a}}.
\newblock URL \url{https://zenodo.org/records/12608602}.

\bibitem[Gao et~al.(2024{\natexlab{b}})Gao, Wettig, Yen, and Chen]{gao2024trainlongcontextlanguagemodels}
Tianyu Gao, Alexander Wettig, Howard Yen, and Danqi Chen.
\newblock How to train long-context language models (effectively), 2024{\natexlab{b}}.
\newblock URL \url{https://arxiv.org/abs/2410.02660}.

\bibitem[Gao et~al.(2024{\natexlab{c}})Gao, Wettig, Yen, and Chen]{gao_how_2024}
Tianyu Gao, Alexander Wettig, Howard Yen, and Danqi Chen.
\newblock How to {Train} {Long}-{Context} {Language} {Models} ({Effectively}), October 2024{\natexlab{c}}.
\newblock URL \url{http://arxiv.org/abs/2410.02660}.
\newblock arXiv:2410.02660.

\bibitem[Goyal et~al.(2023)Goyal, Li, and Durrett]{goyal2023newssummarizationevaluationera}
Tanya Goyal, Junyi~Jessy Li, and Greg Durrett.
\newblock News summarization and evaluation in the era of gpt-3, 2023.
\newblock URL \url{https://arxiv.org/abs/2209.12356}.

\bibitem[Ho et~al.(2023)Ho, Schmid, and Yun]{ho-etal-2023-large}
Namgyu Ho, Laura Schmid, and Se-Young Yun.
\newblock Large language models are reasoning teachers.
\newblock In Anna Rogers, Jordan Boyd-Graber, and Naoaki Okazaki (eds.), \emph{Proceedings of the 61st Annual Meeting of the Association for Computational Linguistics (Volume 1: Long Papers)}, pp.\  14852--14882, Toronto, Canada, July 2023. Association for Computational Linguistics.
\newblock \doi{10.18653/v1/2023.acl-long.830}.
\newblock URL \url{https://aclanthology.org/2023.acl-long.830/}.

\bibitem[Honovich et~al.(2022)Honovich, Shaham, Bowman, and Levy]{honovich2022instructioninductionexamplesnatural}
Or~Honovich, Uri Shaham, Samuel~R. Bowman, and Omer Levy.
\newblock Instruction induction: From few examples to natural language task descriptions, 2022.
\newblock URL \url{https://arxiv.org/abs/2205.10782}.

\bibitem[Hsieh et~al.(2023)Hsieh, Li, Yeh, Nakhost, Fujii, Ratner, Krishna, Lee, and Pfister]{hsieh-etal-2023-distilling}
Cheng-Yu Hsieh, Chun-Liang Li, Chih-kuan Yeh, Hootan Nakhost, Yasuhisa Fujii, Alex Ratner, Ranjay Krishna, Chen-Yu Lee, and Tomas Pfister.
\newblock Distilling step-by-step! outperforming larger language models with less training data and smaller model sizes.
\newblock In Anna Rogers, Jordan Boyd-Graber, and Naoaki Okazaki (eds.), \emph{Findings of the Association for Computational Linguistics: ACL 2023}, pp.\  8003--8017, Toronto, Canada, July 2023. Association for Computational Linguistics.
\newblock \doi{10.18653/v1/2023.findings-acl.507}.
\newblock URL \url{https://aclanthology.org/2023.findings-acl.507/}.

\bibitem[Huang et~al.(2023)Huang, Gu, Hou, Wu, Wang, Yu, and Han]{huang-etal-2023-large}
Jiaxin Huang, Shixiang Gu, Le~Hou, Yuexin Wu, Xuezhi Wang, Hongkun Yu, and Jiawei Han.
\newblock Large language models can self-improve.
\newblock In Houda Bouamor, Juan Pino, and Kalika Bali (eds.), \emph{Proceedings of the 2023 Conference on Empirical Methods in Natural Language Processing}, pp.\  1051--1068, Singapore, December 2023. Association for Computational Linguistics.
\newblock \doi{10.18653/v1/2023.emnlp-main.67}.
\newblock URL \url{https://aclanthology.org/2023.emnlp-main.67/}.

\bibitem[Jacobs et~al.(2023)Jacobs, Tanaka, Zhang, Zhang, Song, Rajbhandari, and He]{jacobs_deepspeed_2023}
Sam~Ade Jacobs, Masahiro Tanaka, Chengming Zhang, Minjia Zhang, Shuaiwen~Leon Song, Samyam Rajbhandari, and Yuxiong He.
\newblock {DeepSpeed} {Ulysses}: {System} {Optimizations} for {Enabling} {Training} of {Extreme} {Long} {Sequence} {Transformer} {Models}, October 2023.
\newblock URL \url{http://arxiv.org/abs/2309.14509}.
\newblock arXiv:2309.14509 [cs].

\bibitem[Karpinska et~al.(2024)Karpinska, Thai, Lo, Goyal, and Iyyer]{karpinska_one_2024}
Marzena Karpinska, Katherine Thai, Kyle Lo, Tanya Goyal, and Mohit Iyyer.
\newblock One {Thousand} and {One} {Pairs}: {A} "novel" challenge for long-context language models, July 2024.
\newblock URL \url{http://arxiv.org/abs/2406.16264}.
\newblock arXiv:2406.16264 [cs].

\bibitem[Kim et~al.(2024)Kim, Chang, Karpinska, Garimella, Manjunatha, Lo, Goyal, and Iyyer]{kim_fables_2024}
Yekyung Kim, Yapei Chang, Marzena Karpinska, Aparna Garimella, Varun Manjunatha, Kyle Lo, Tanya Goyal, and Mohit Iyyer.
\newblock {FABLES}: {Evaluating} faithfulness and content selection in book-length summarization, April 2024.
\newblock URL \url{http://arxiv.org/abs/2404.01261}.
\newblock arXiv:2404.01261 [cs].

\bibitem[Kojima et~al.(2023)Kojima, Gu, Reid, Matsuo, and Iwasawa]{kojima2023largelanguagemodelszeroshot}
Takeshi Kojima, Shixiang~Shane Gu, Machel Reid, Yutaka Matsuo, and Yusuke Iwasawa.
\newblock Large language models are zero-shot reasoners, 2023.
\newblock URL \url{https://arxiv.org/abs/2205.11916}.

\bibitem[Kočiský et~al.(2018)Kočiský, Schwarz, Blunsom, Dyer, Hermann, Melis, and Grefenstette]{kocisky_narrativeqa_2018}
Tomáš Kočiský, Jonathan Schwarz, Phil Blunsom, Chris Dyer, Karl~Moritz Hermann, Gábor Melis, and Edward Grefenstette.
\newblock The {NarrativeQA} {Reading} {Comprehension} {Challenge}.
\newblock \emph{Transactions of the Association for Computational Linguistics}, 6:\penalty0 317--328, 2018.
\newblock \doi{10.1162/tacl_a_00023}.
\newblock URL \url{https://aclanthology.org/Q18-1023}.
\newblock Place: Cambridge, MA Publisher: MIT Press.

\bibitem[Köksal et~al.(2023)Köksal, Schick, Korhonen, and Schütze]{koksal_longform_2023}
Abdullatif Köksal, Timo Schick, Anna Korhonen, and Hinrich Schütze.
\newblock {LongForm}: {Optimizing} {Instruction} {Tuning} for {Long} {Text} {Generation} with {Corpus} {Extraction}, April 2023.
\newblock URL \url{http://arxiv.org/abs/2304.08460}.
\newblock arXiv:2304.08460 [cs].

\bibitem[Lambert et~al.(2024)Lambert, Morrison, Pyatkin, Huang, Ivison, Brahman, Miranda, Liu, Dziri, Lyu, Gu, Malik, Graf, Hwang, Yang, Bras, Tafjord, Wilhelm, Soldaini, Smith, Wang, Dasigi, and Hajishirzi]{lambert2024tulu3}
Nathan Lambert, Jacob Morrison, Valentina Pyatkin, Shengyi Huang, Hamish Ivison, Faeze Brahman, Lester James~V. Miranda, Alisa Liu, Nouha Dziri, Shane Lyu, Yuling Gu, Saumya Malik, Victoria Graf, Jena~D. Hwang, Jiangjiang Yang, Ronan~Le Bras, Oyvind Tafjord, Chris Wilhelm, Luca Soldaini, Noah~A. Smith, Yizhong Wang, Pradeep Dasigi, and Hannaneh Hajishirzi.
\newblock Tülu 3: Pushing frontiers in open language model post-training, 2024.

\bibitem[Levy et~al.(2024)Levy, Jacoby, and Goldberg]{levy_same_2024}
Mosh Levy, Alon Jacoby, and Yoav Goldberg.
\newblock Same {Task}, {More} {Tokens}: the {Impact} of {Input} {Length} on the {Reasoning} {Performance} of {Large} {Language} {Models}, February 2024.
\newblock URL \url{http://arxiv.org/abs/2402.14848}.
\newblock arXiv:2402.14848 [cs].

\bibitem[Li et~al.(2025)Li, Sun, Huang, Zhong, Jiang, Han, Zhang, Wang, and Liu]{li2025preferenceleakagecontaminationproblem}
Dawei Li, Renliang Sun, Yue Huang, Ming Zhong, Bohan Jiang, Jiawei Han, Xiangliang Zhang, Wei Wang, and Huan Liu.
\newblock Preference leakage: A contamination problem in llm-as-a-judge, 2025.
\newblock URL \url{https://arxiv.org/abs/2502.01534}.

\bibitem[Li et~al.(2023)Li, Hessel, Yu, Ren, Chang, and Choi]{li-etal-2023-symbolic}
Liunian~Harold Li, Jack Hessel, Youngjae Yu, Xiang Ren, Kai-Wei Chang, and Yejin Choi.
\newblock Symbolic chain-of-thought distillation: Small models can also {\textquotedblleft}think{\textquotedblright} step-by-step.
\newblock In Anna Rogers, Jordan Boyd-Graber, and Naoaki Okazaki (eds.), \emph{Proceedings of the 61st Annual Meeting of the Association for Computational Linguistics (Volume 1: Long Papers)}, pp.\  2665--2679, Toronto, Canada, July 2023. Association for Computational Linguistics.
\newblock \doi{10.18653/v1/2023.acl-long.150}.
\newblock URL \url{https://aclanthology.org/2023.acl-long.150/}.

\bibitem[Li et~al.(2024)Li, Yu, Zhou, Schick, Levy, Zettlemoyer, Weston, and Lewis]{li2024selfalignmentinstructionbacktranslation}
Xian Li, Ping Yu, Chunting Zhou, Timo Schick, Omer Levy, Luke Zettlemoyer, Jason Weston, and Mike Lewis.
\newblock Self-alignment with instruction backtranslation, 2024.
\newblock URL \url{https://arxiv.org/abs/2308.06259}.

\bibitem[Lieber et~al.(2024)Lieber, Lenz, Bata, Cohen, Osin, Dalmedigos, Safahi, Meirom, Belinkov, Shalev-Shwartz, Abend, Alon, Asida, Bergman, Glozman, Gokhman, Manevich, Ratner, Rozen, Shwartz, Zusman, and Shoham]{lieber2024jambahybridtransformermambalanguage}
Opher Lieber, Barak Lenz, Hofit Bata, Gal Cohen, Jhonathan Osin, Itay Dalmedigos, Erez Safahi, Shaked Meirom, Yonatan Belinkov, Shai Shalev-Shwartz, Omri Abend, Raz Alon, Tomer Asida, Amir Bergman, Roman Glozman, Michael Gokhman, Avashalom Manevich, Nir Ratner, Noam Rozen, Erez Shwartz, Mor Zusman, and Yoav Shoham.
\newblock Jamba: A hybrid transformer-mamba language model, 2024.
\newblock URL \url{https://arxiv.org/abs/2403.19887}.

\bibitem[Lin(2004)]{lin-2004-rouge}
Chin-Yew Lin.
\newblock {ROUGE}: A package for automatic evaluation of summaries.
\newblock In \emph{Text Summarization Branches Out}, pp.\  74--81, Barcelona, Spain, July 2004. Association for Computational Linguistics.
\newblock URL \url{https://aclanthology.org/W04-1013/}.

\bibitem[Liu \& Abbeel(2023)Liu and Abbeel]{liu2023blockwise}
Hao Liu and Pieter Abbeel.
\newblock Blockwise parallel transformer for large context models.
\newblock \emph{Advances in neural information processing systems}, 2023.

\bibitem[Liu et~al.(2023)Liu, Zaharia, and Abbeel]{liu2023ring}
Hao Liu, Matei Zaharia, and Pieter Abbeel.
\newblock Ring attention with blockwise transformers for near-infinite context.
\newblock \emph{arXiv preprint arXiv:2310.01889}, 2023.

\bibitem[Liu et~al.(2024)Liu, Lin, Hewitt, Paranjape, Bevilacqua, Petroni, and Liang]{liu2024lost}
Nelson~F Liu, Kevin Lin, John Hewitt, Ashwin Paranjape, Michele Bevilacqua, Fabio Petroni, and Percy Liang.
\newblock Lost in the middle: How language models use long contexts.
\newblock \emph{Transactions of the Association for Computational Linguistics}, 12:\penalty0 157--173, 2024.

\bibitem[McNemar(1947)]{mcnemar1947note}
Quinn McNemar.
\newblock Note on the sampling error of the difference between correlated proportions or percentages.
\newblock \emph{Psychometrika}, 12\penalty0 (2):\penalty0 153--157, 1947.

\bibitem[Muennighoff et~al.(2025)Muennighoff, Yang, Shi, Li, Fei-Fei, Hajishirzi, Zettlemoyer, Liang, Candès, and Hashimoto]{muennighoff2025s1simpletesttimescaling}
Niklas Muennighoff, Zitong Yang, Weijia Shi, Xiang~Lisa Li, Li~Fei-Fei, Hannaneh Hajishirzi, Luke Zettlemoyer, Percy Liang, Emmanuel Candès, and Tatsunori Hashimoto.
\newblock s1: Simple test-time scaling, 2025.
\newblock URL \url{https://arxiv.org/abs/2501.19393}.

\bibitem[{OpenAI}(2024)]{openai_o1_2024}
{OpenAI}.
\newblock o1 {System} {Card}, December 2024.
\newblock URL \url{https://cdn.openai.com/o1-system-card-20241205.pdf}.

\bibitem[OpenAI et~al.(2024)OpenAI, Achiam, Adler, Agarwal, Ahmad, Akkaya, Aleman, Almeida, Altenschmidt, Altman, et~al.]{openai2024gpt4technicalreport}
OpenAI, Josh Achiam, Steven Adler, Sandhini Agarwal, Lama Ahmad, Ilge Akkaya, Florencia~Leoni Aleman, Diogo Almeida, Janko Altenschmidt, Sam Altman, and 271 others.
\newblock Gpt-4 technical report, 2024.
\newblock URL \url{https://arxiv.org/abs/2303.08774}.

\bibitem[Panickssery et~al.(2024)Panickssery, Bowman, and Feng]{panickssery2024llmevaluatorsrecognizefavor}
Arjun Panickssery, Samuel~R. Bowman, and Shi Feng.
\newblock Llm evaluators recognize and favor their own generations, 2024.
\newblock URL \url{https://arxiv.org/abs/2404.13076}.

\bibitem[Paszke et~al.(2019)Paszke, Gross, Massa, Lerer, Bradbury, Chanan, Killeen, Lin, Gimelshein, Antiga, Desmaison, Köpf, Yang, DeVito, Raison, Tejani, Chilamkurthy, Steiner, Fang, Bai, and Chintala]{paszke2019pytorchimperativestylehighperformance}
Adam Paszke, Sam Gross, Francisco Massa, Adam Lerer, James Bradbury, Gregory Chanan, Trevor Killeen, Zeming Lin, Natalia Gimelshein, Luca Antiga, Alban Desmaison, Andreas Köpf, Edward Yang, Zach DeVito, Martin Raison, Alykhan Tejani, Sasank Chilamkurthy, Benoit Steiner, Lu~Fang, Junjie Bai, and Soumith Chintala.
\newblock Pytorch: An imperative style, high-performance deep learning library, 2019.
\newblock URL \url{https://arxiv.org/abs/1912.01703}.

\bibitem[Peng et~al.(2023)Peng, Quesnelle, Fan, and Shippole]{peng_yarn_2023}
Bowen Peng, Jeffrey Quesnelle, Honglu Fan, and Enrico Shippole.
\newblock {YaRN}: {Efficient} {Context} {Window} {Extension} of {Large} {Language} {Models}, November 2023.
\newblock URL \url{http://arxiv.org/abs/2309.00071}.
\newblock arXiv:2309.00071 [cs].

\bibitem[Pham et~al.(2024)Pham, Sun, and Iyyer]{pham_suri_2024}
Chau~Minh Pham, Simeng Sun, and Mohit Iyyer.
\newblock Suri: {Multi}-constraint {Instruction} {Following} for {Long}-form {Text} {Generation}, June 2024.
\newblock URL \url{http://arxiv.org/abs/2406.19371}.
\newblock arXiv:2406.19371 [cs].

\bibitem[Press et~al.(2022)Press, Smith, and Lewis]{press_train_2022}
Ofir Press, Noah~A. Smith, and Mike Lewis.
\newblock Train {Short}, {Test} {Long}: {Attention} with {Linear} {Biases} {Enables} {Input} {Length} {Extrapolation}, April 2022.
\newblock URL \url{http://arxiv.org/abs/2108.12409}.
\newblock arXiv:2108.12409 [cs].

\bibitem[Puerto et~al.(2024)Puerto, Chubakov, Zhu, Madabushi, and Gurevych]{puerto2024finetuningdivergentchainsthought}
Haritz Puerto, Tilek Chubakov, Xiaodan Zhu, Harish~Tayyar Madabushi, and Iryna Gurevych.
\newblock Fine-tuning with divergent chains of thought boosts reasoning through self-correction in language models, 2024.
\newblock URL \url{https://arxiv.org/abs/2407.03181}.

\bibitem[Qi et~al.(2024)Qi, Luo, Huang, Zhao, Jiang, Fan, Lakkaraju, and Glass]{qi2024quantifyinggeneralizationcomplexitylarge}
Zhenting Qi, Hongyin Luo, Xuliang Huang, Zhuokai Zhao, Yibo Jiang, Xiangjun Fan, Himabindu Lakkaraju, and James Glass.
\newblock Quantifying generalization complexity for large language models, 2024.
\newblock URL \url{https://arxiv.org/abs/2410.01769}.

\bibitem[Qwen et~al.(2024)Qwen, Yang, Yang, Zhang, Hui, Zheng, Yu, Li, Liu, Huang, Wei, Lin, Yang, Tu, Zhang, Yang, Yang, Zhou, Lin, Dang, Lu, Bao, Yang, Yu, Li, Xue, Zhang, Zhu, Men, Lin, Li, Xia, Ren, Ren, Fan, Su, Zhang, Wan, Liu, Cui, Zhang, and Qiu]{qwen_qwen25_2024}
Qwen, An~Yang, Baosong Yang, Beichen Zhang, Binyuan Hui, Bo~Zheng, Bowen Yu, Chengyuan Li, Dayiheng Liu, Fei Huang, Haoran Wei, Huan Lin, Jian Yang, Jianhong Tu, Jianwei Zhang, Jianxin Yang, Jiaxi Yang, Jingren Zhou, Junyang Lin, Kai Dang, Keming Lu, Keqin Bao, Kexin Yang, Le~Yu, Mei Li, Mingfeng Xue, Pei Zhang, Qin Zhu, Rui Men, Runji Lin, Tianhao Li, Tingyu Xia, Xingzhang Ren, Xuancheng Ren, Yang Fan, Yang Su, Yichang Zhang, Yu~Wan, Yuqiong Liu, Zeyu Cui, Zhenru Zhang, and Zihan Qiu.
\newblock Qwen2.5 {Technical} {Report}, December 2024.
\newblock URL \url{http://arxiv.org/abs/2412.15115}.
\newblock arXiv:2412.15115 [cs].

\bibitem[Robertson et~al.(1995)Robertson, Walker, Jones, Hancock-Beaulieu, Gatford, et~al.]{robertson1995okapi}
Stephen~E Robertson, Steve Walker, Susan Jones, Micheline~M Hancock-Beaulieu, Mike Gatford, et~al.
\newblock \emph{Okapi at TREC-3}.
\newblock British Library Research and Development Department, 1995.

\bibitem[Sprague et~al.(2024)Sprague, Ye, Bostrom, Chaudhuri, and Durrett]{sprague_musr_2024}
Zayne Sprague, Xi~Ye, Kaj Bostrom, Swarat Chaudhuri, and Greg Durrett.
\newblock {MuSR}: {Testing} the {Limits} of {Chain}-of-thought with {Multistep} {Soft} {Reasoning}, March 2024.
\newblock URL \url{http://arxiv.org/abs/2310.16049}.
\newblock arXiv:2310.16049 [cs].

\bibitem[Su et~al.(2023)Su, Lu, Pan, Murtadha, Wen, and Liu]{su_roformer_2023}
Jianlin Su, Yu~Lu, Shengfeng Pan, Ahmed Murtadha, Bo~Wen, and Yunfeng Liu.
\newblock {RoFormer}: {Enhanced} {Transformer} with {Rotary} {Position} {Embedding}, November 2023.
\newblock URL \url{http://arxiv.org/abs/2104.09864}.
\newblock arXiv:2104.09864 [cs].

\bibitem[Team et~al.(2024)Team, Georgiev, Lei, Burnell, Bai, Gulati, Tanzer, Vincent, Pan, Wang, et~al.]{geminiteam2024gemini15unlockingmultimodal}
Gemini Team, Petko Georgiev, Ving~Ian Lei, Ryan Burnell, Libin Bai, Anmol Gulati, Garrett Tanzer, Damien Vincent, Zhufeng Pan, Shibo Wang, and 1127 others.
\newblock Gemini 1.5: Unlocking multimodal understanding across millions of tokens of context, 2024.
\newblock URL \url{https://arxiv.org/abs/2403.05530}.

\bibitem[Team(2024)]{qwen2.5}
Qwen Team.
\newblock Qwen2.5: A party of foundation models, September 2024.
\newblock URL \url{https://qwenlm.github.io/blog/qwen2.5/}.

\bibitem[Wang et~al.(2024)Wang, Yang, Zhang, Huang, and Wei]{wang2024bootstrapcontextlength}
Liang Wang, Nan Yang, Xingxing Zhang, Xiaolong Huang, and Furu Wei.
\newblock Bootstrap your own context length, 2024.
\newblock URL \url{https://arxiv.org/abs/2412.18860}.

\bibitem[Wang et~al.(2023{\natexlab{a}})Wang, Wei, Schuurmans, Le, Chi, Narang, Chowdhery, and Zhou]{wang2023selfconsistencyimproveschainthought}
Xuezhi Wang, Jason Wei, Dale Schuurmans, Quoc Le, Ed~Chi, Sharan Narang, Aakanksha Chowdhery, and Denny Zhou.
\newblock Self-consistency improves chain of thought reasoning in language models, 2023{\natexlab{a}}.
\newblock URL \url{https://arxiv.org/abs/2203.11171}.

\bibitem[Wang et~al.(2023{\natexlab{b}})Wang, Kordi, Mishra, Liu, Smith, Khashabi, and Hajishirzi]{wang2023selfinstructaligninglanguagemodels}
Yizhong Wang, Yeganeh Kordi, Swaroop Mishra, Alisa Liu, Noah~A. Smith, Daniel Khashabi, and Hannaneh Hajishirzi.
\newblock Self-instruct: Aligning language models with self-generated instructions, 2023{\natexlab{b}}.
\newblock URL \url{https://arxiv.org/abs/2212.10560}.

\bibitem[Wei et~al.(2023)Wei, Wang, Schuurmans, Bosma, Ichter, Xia, Chi, Le, and Zhou]{wei_chain--thought_2023}
Jason Wei, Xuezhi Wang, Dale Schuurmans, Maarten Bosma, Brian Ichter, Fei Xia, Ed~Chi, Quoc Le, and Denny Zhou.
\newblock Chain-of-{Thought} {Prompting} {Elicits} {Reasoning} in {Large} {Language} {Models}, January 2023.
\newblock URL \url{http://arxiv.org/abs/2201.11903}.
\newblock arXiv:2201.11903.

\bibitem[Wilcoxon(1992)]{wilcoxon1992individual}
Frank Wilcoxon.
\newblock Individual comparisons by ranking methods.
\newblock In \emph{Breakthroughs in statistics: Methodology and distribution}, pp.\  196--202. Springer, 1992.

\bibitem[Wolf et~al.(2020)Wolf, Debut, Sanh, Chaumond, Delangue, Moi, Cistac, Rault, Louf, Funtowicz, Davison, Shleifer, von Platen, Ma, Jernite, Plu, Xu, Scao, Gugger, Drame, Lhoest, and Rush]{wolf2020huggingfacestransformersstateoftheartnatural}
Thomas Wolf, Lysandre Debut, Victor Sanh, Julien Chaumond, Clement Delangue, Anthony Moi, Pierric Cistac, Tim Rault, Rémi Louf, Morgan Funtowicz, Joe Davison, Sam Shleifer, Patrick von Platen, Clara Ma, Yacine Jernite, Julien Plu, Canwen Xu, Teven~Le Scao, Sylvain Gugger, Mariama Drame, Quentin Lhoest, and Alexander~M. Rush.
\newblock Huggingface's transformers: State-of-the-art natural language processing, 2020.
\newblock URL \url{https://arxiv.org/abs/1910.03771}.

\bibitem[Wu et~al.(2024{\natexlab{a}})Wu, Wang, Fu, Yue, Zhu, and Li]{wu2024longcontextalignmentshort}
Wenhao Wu, Yizhong Wang, Yao Fu, Xiang Yue, Dawei Zhu, and Sujian Li.
\newblock Long context alignment with short instructions and synthesized positions, 2024{\natexlab{a}}.
\newblock URL \url{https://arxiv.org/abs/2405.03939}.

\bibitem[Wu et~al.(2024{\natexlab{b}})Wu, Wang, Liu, Shi, Yan, Lu, Zhu, and Zhang]{wu_lifbench_2024}
Xiaodong Wu, Minhao Wang, Yichen Liu, Xiaoming Shi, He~Yan, Xiangju Lu, Junmin Zhu, and Wei Zhang.
\newblock {LIFBench}: {Evaluating} the {Instruction} {Following} {Performance} and {Stability} of {Large} {Language} {Models} in {Long}-{Context} {Scenarios}, November 2024{\natexlab{b}}.
\newblock URL \url{http://arxiv.org/abs/2411.07037}.
\newblock arXiv:2411.07037.

\bibitem[Xiong et~al.(2023)Xiong, Liu, Molybog, Zhang, Bhargava, Hou, Martin, Rungta, Sankararaman, Oguz, Khabsa, Fang, Mehdad, Narang, Malik, Fan, Bhosale, Edunov, Lewis, Wang, and Ma]{xiong_effective_2023}
Wenhan Xiong, Jingyu Liu, Igor Molybog, Hejia Zhang, Prajjwal Bhargava, Rui Hou, Louis Martin, Rashi Rungta, Karthik~Abinav Sankararaman, Barlas Oguz, Madian Khabsa, Han Fang, Yashar Mehdad, Sharan Narang, Kshitiz Malik, Angela Fan, Shruti Bhosale, Sergey Edunov, Mike Lewis, Sinong Wang, and Hao Ma.
\newblock Effective {Long}-{Context} {Scaling} of {Foundation} {Models}, November 2023.
\newblock URL \url{http://arxiv.org/abs/2309.16039}.
\newblock arXiv:2309.16039 [cs].

\bibitem[Xiong et~al.(2024)Xiong, Papageorgiou, Lee, and Papailiopoulos]{xiong2024artificialneedlesrealhaystacks}
Zheyang Xiong, Vasilis Papageorgiou, Kangwook Lee, and Dimitris Papailiopoulos.
\newblock From artificial needles to real haystacks: Improving retrieval capabilities in llms by finetuning on synthetic data, 2024.
\newblock URL \url{https://arxiv.org/abs/2406.19292}.

\bibitem[Xu et~al.(2024{\natexlab{a}})Xu, Ping, Wu, Xu, Liu, Shoeybi, and Catanzaro]{xu2024chatqa2bridginggap}
Peng Xu, Wei Ping, Xianchao Wu, Chejian Xu, Zihan Liu, Mohammad Shoeybi, and Bryan Catanzaro.
\newblock Chatqa 2: Bridging the gap to proprietary llms in long context and rag capabilities, 2024{\natexlab{a}}.
\newblock URL \url{https://arxiv.org/abs/2407.14482}.

\bibitem[Xu et~al.(2024{\natexlab{b}})Xu, Zhu, Zhao, Pan, Li, and Wang]{xu2024prideprejudicellmamplifies}
Wenda Xu, Guanglei Zhu, Xuandong Zhao, Liangming Pan, Lei Li, and William~Yang Wang.
\newblock Pride and prejudice: Llm amplifies self-bias in self-refinement, 2024{\natexlab{b}}.
\newblock URL \url{https://arxiv.org/abs/2402.11436}.

\bibitem[Yang et~al.(2025)Yang, Yu, Li, Liu, Huang, Huang, Tu, Zhang, Zhou, Lin, Dang, Yang, Li, Sun, Zhu, Men, He, Xu, Yin, Qiu, Ren, Yang, Li, Xu, and Zhang]{yang_qwen25-1m_2025}
An~Yang, Bowen Yu, Chengyuan Li, Dayiheng Liu, Fei Huang, Haoyan Huang, Jianhong Tu, Jianwei Zhang, Jingren Zhou, Junyang Lin, Kai Dang, Kexin Yang, Mei Li, Minmin Sun, Qin Zhu, Rui Men, Tao He, Weijia Xu, Wenbiao Yin, Xiafei Qiu, Xingzhang Ren, Xinlong Yang, Yong Li, Zhiying Xu, and Zipeng Zhang.
\newblock Qwen2.5-{1M} {Technical} {Report}, January 2025.

\bibitem[Yao et~al.(2023)Yao, Yu, Zhao, Shafran, Griffiths, Cao, and Narasimhan]{yao2023treethoughtsdeliberateproblem}
Shunyu Yao, Dian Yu, Jeffrey Zhao, Izhak Shafran, Thomas~L. Griffiths, Yuan Cao, and Karthik Narasimhan.
\newblock Tree of thoughts: Deliberate problem solving with large language models, 2023.
\newblock URL \url{https://arxiv.org/abs/2305.10601}.

\bibitem[Yen et~al.(2024)Yen, Gao, Hou, Ding, Fleischer, Izsak, Wasserblat, and Chen]{yen2024helmet}
Howard Yen, Tianyu Gao, Minmin Hou, Ke~Ding, Daniel Fleischer, Peter Izsak, Moshe Wasserblat, and Danqi Chen.
\newblock Helmet: How to evaluate long-context language models effectively and thoroughly.
\newblock \emph{arXiv preprint arXiv:2410.02694}, 2024.

\bibitem[Yeo et~al.(2025)Yeo, Tong, Niu, Neubig, and Yue]{yeo2025demystifyinglongchainofthoughtreasoning}
Edward Yeo, Yuxuan Tong, Morry Niu, Graham Neubig, and Xiang Yue.
\newblock Demystifying long chain-of-thought reasoning in llms, 2025.
\newblock URL \url{https://arxiv.org/abs/2502.03373}.

\bibitem[Zelikman et~al.(2022)Zelikman, Wu, Mu, and Goodman]{zelikman2022starbootstrappingreasoningreasoning}
Eric Zelikman, Yuhuai Wu, Jesse Mu, and Noah~D. Goodman.
\newblock Star: Bootstrapping reasoning with reasoning, 2022.
\newblock URL \url{https://arxiv.org/abs/2203.14465}.

\bibitem[Zhang et~al.(2024)Zhang, Chen, Hu, Xu, Chen, Hao, Han, Thai, Wang, Liu, and Sun]{zhang2024inftybenchextendinglongcontext}
Xinrong Zhang, Yingfa Chen, Shengding Hu, Zihang Xu, Junhao Chen, Moo~Khai Hao, Xu~Han, Zhen~Leng Thai, Shuo Wang, Zhiyuan Liu, and Maosong Sun.
\newblock $\infty$bench: Extending long context evaluation beyond 100k tokens, 2024.
\newblock URL \url{https://arxiv.org/abs/2402.13718}.

\bibitem[Zhou et~al.(2023)Zhou, Muresanu, Han, Paster, Pitis, Chan, and Ba]{zhou2023largelanguagemodelshumanlevel}
Yongchao Zhou, Andrei~Ioan Muresanu, Ziwen Han, Keiran Paster, Silviu Pitis, Harris Chan, and Jimmy Ba.
\newblock Large language models are human-level prompt engineers, 2023.
\newblock URL \url{https://arxiv.org/abs/2211.01910}.

\end{thebibliography}
